\documentclass[twoside,11pt]{article}
\pdfoutput=1
\usepackage{jmlr2e}
\usepackage{amsmath}
\usepackage{subfigure}
\usepackage{color}
\usepackage{algorithm}
\usepackage{clrscode}
\usepackage{url}

\def\DP{\mathop{\mathrm{DP}}}
\def\Dir{\mathop{\mathrm{Dir}}}

\def\g{\Gamma}

\def\C{\mathrm{C}}
\def\alphadp{\alpha_{\mathit{DP}}}
\def\alphadir{\alpha_{\mathit{Dir}}}
\def\lset{{\cal L}}\def\iset{{\cal I}}
\def\jset{{\cal J}}
\def\zset{{\cal Z}}

\firstpageno{1}

\begin{document}

\title{Component models for large networks}

\author{\name Janne Sinkkonen \email janne.sinkkonen@tkk.fi \\   
       \addr 
       Xtract Ltd. and Helsinki University of Technology
       \AND
       \name Janne Aukia \email janne.aukia@xtract.com \\ 
       \addr
              Xtract Ltd.\\ Hitsaajankatu 22, 00810 Helsinki, Finland
       \AND
       \name Samuel Kaski \email samuel.kaski@tkk.fi \\    
        \addr Department of Information and Computer Science\\
        \addr Helsinki University of Technology\\ P.O. Box 5400, FI-02015 TKK, Finland}

\editor{NN}

\maketitle

\begin{abstract}%   <- trailing '%' for backward compatibility of .sty file
  Being among the easiest ways to find meaningful structure from
  discrete data, Latent Dirichlet Allocation (LDA) and related
  component models have been applied widely. They are simple,
  computationally fast and scalable, interpretable, and admit
  nonparametric priors. In the currently popular field of network
  modeling, relatively little work has taken uncertainty of data
  seriously in the Bayesian sense, and component models have been
  introduced to the field only recently, by treating each node as a
  bag of out-going links. We introduce an alternative, interaction
  component model for communities (ICMc), where the whole network is a
  bag of links, stemming from different components. The former finds
  both disassortative and assortative structure, while the alternative
  assumes assortativity and finds community-like structures like the
  earlier methods motivated by physics. With Dirichlet Process priors
  and an efficient implementation the models are highly scalable, as
  demonstrated with a social network from the Last.fm web site, with
  670,000 nodes and 1.89 million links.
\end{abstract}


% \subsection*{Notation (to be removed)}

% [Notation; examples do not yet include LDA: Functions and operators by
% roman letters, with an exception: normalizer $\mathrm{Z}(\alphadir,
% \alphadp,\beta)$, Gamma $\g$, the chooser $\mathrm{C}(n, \alpha)$,
% complexity $\O()$. Complex objects or exchangeable sets by
% calligraphic letters: link data $\lset$ and the associated hidden
% variables $\zset$, link endpoints ${\iset}$ and ${\jset}$. Matrices
% and vectors by small roman letters, with an exception: Dirichlet
% sample for components $\theta$, link counts over components $n$, link
% endpoint counts over components and nodes $k$. Constant integers by
% capitalized letters: $K$ the number of components, $L$ the number of
% links, $I$ the number of nodes, $G$ the number of Gibbs iterations
% (one goes over the whole data set). Scalars by Greek letters: $\alpha,
% \beta$.  Average degree is $\rho$. Indices: components $z$, nodes $i$,
% links $l$. Single link $l_0$, with the associated latent variable
% $z_0$. Dotted notation is mainly for Gibbs counts with a sample (link)
% removed. We have nodes and links, no vertices and edges.]
\section{Introduction} % SK

% Motivation, context, intro to our work etc.
% Sparse, large, and implications for model architecture

% Related:
% Zhang et al., Airoldi, Newman, modularity
% Secondary: Handcock07 (logistic latent space model), Daudin07

\begin{figure}
\centerline{
  \includegraphics[width=0.6\textwidth]{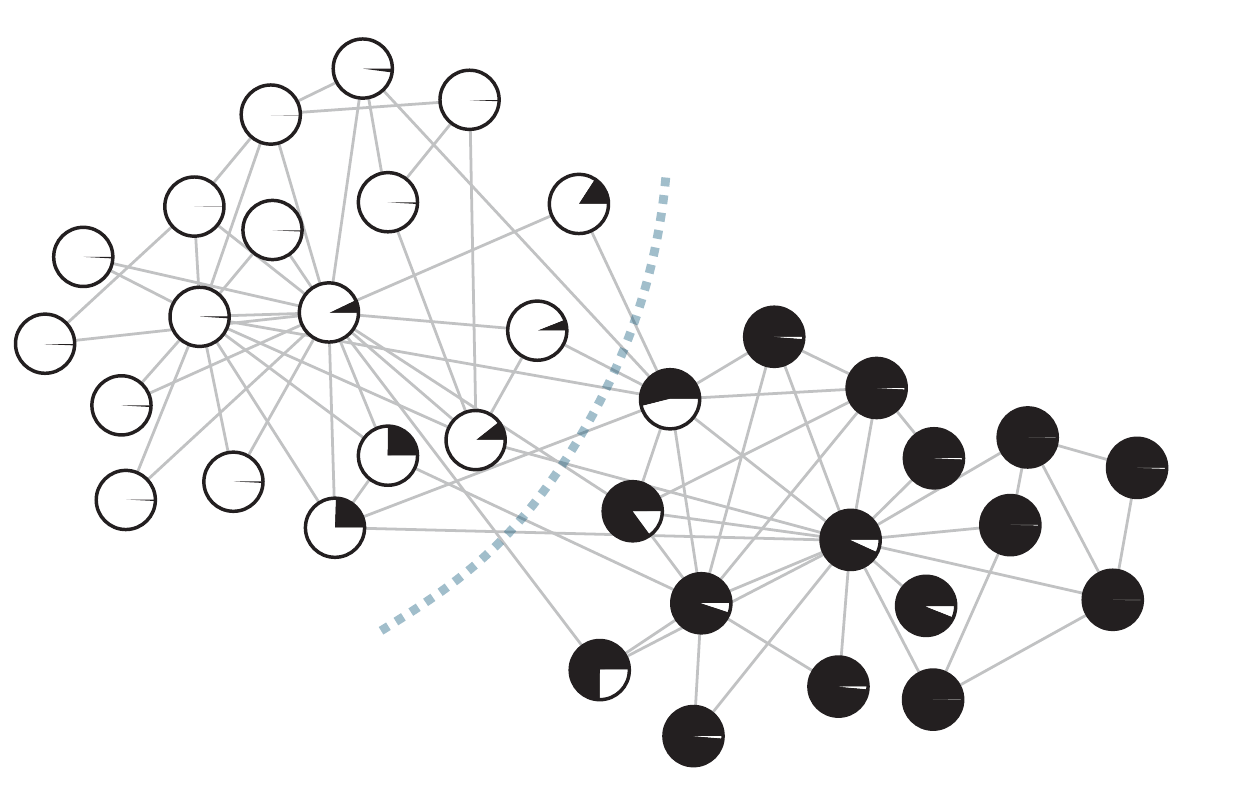}
}
\caption{One of the models (ICMc) on a classic small network, the
  Karate club. \citet{Zachary77} observed the social interactions
  (edges) between 34 members of a karate club over two years. During
  this period, there was disagreement among the club members which
  eventually led to the splitting of the club (dashed line).  Black
  and white depict the degree of component memberships obtained by the
  model, without knowing the correct split.}
\label{fig:demo}
\end{figure}

Data collections representable as networks, or sets of binary
relations between vertices, appear now frequently in many fields,
including social networks and interaction networks in biology
(Fig.~\ref{fig:demo}). Consequently, inferring properties of the
network vertices\footnote{We will use the terms vertex and node
  interchangeably, and likewise for edges and links.}  from the edges
has become a common data mining problem. Most of the work has been
about dividing the vertices into relatively well-connected subsets, or
\emph{communities} \citep{Fortunato07r}. Most papers on communities
have been inspired by graph theory and physics, as is a large field of
fundamental network-related work not directly relevant here.
Especially optimizing a measure of good division called
\emph{modularity} \citep{Newman06a} has gained success but is not
without its problems \citep{Fortunato07,Kumpula07}.

A feature and potential problem of modularity is that it takes the
observed edges granted, while network data are typically not a
complete description of reality but comes with errors, omissions and
uncertainties. Some links may be spurious, for instance due to
measurement noise in biological networks, and some potential links may
be missing, for instance friendship links of newcomers in social
networks. Probabilistic generative models are a tool for modeling and
inference under such uncertainty. They treat the links as random
events, and give an explicit structure for the observed data and its
uncertainty. Compared to non-stochastic methods, they are therefore
likely to perform well as long as their assumptions are valid: They
may reveal properties of networks that are difficult to observe with
non-statistical techniques from the noisy and incomplete data, and
they also offer a groundwork for new conceptual developments. For
example, it may be argued that network communities should be defined
in terms of stochastic models that do not take links at face value but
instead give them an underlying stochastic structure that should be
realistic given an application. On the down side, probabilistic
methods are not always scalable, and they may be difficult to
understand, apply and trust by people from other fields, especially if
the estimation process is complex.

Probabilistic models of network connectivity have been introduced
recently. Mixtures of latent components \citep{Newmanleicht06},
analogous to finite mixture models for vectorial data, are attractive
because of ease of interpretation, but the extensive numbers of
parameters encumber straightforward fitting attempts.  A very
promising development called stochastic block models
(\citealp{Airoldi08}---but also \citealp{Daudin07,Hofman07}) groups
the nodes into blocks and explains the links in terms of homogeneous
connections between pairs of groups. Finally, links can be explained
by the proximity of nodes in a latent space created by a logistic link
\citep{Handcock07}. These models have been successively applied to
various networks from sociology and biology, up to the size of
thousands or tens of thousands of nodes. With heuristic improvements,
stochastic block models are expected to scale up to over one million
nodes (E. Airoldi, p.c.), but in general the computational bottleneck
is scalability.
 
The models discussed in this paper are generative probabilistic models
that decompose the links into components, but their structure makes
them scalable to networks with at least $10^3 \dots 10^6$ nodes, and
up to thousands of latent components---as long as the networks are
sparse enough.  The Simple Social Network LDA (SSN-LDA) model
presented by \citet{Zhang07isi} is identical to the Latent Dirichlet
Allocation (LDA; \citealp{Buntine02,Blei03}) model, originally applied
to text collections. It is also a conceptual although not a geneologic
successor of the mixture model by \citet{Newmanleicht06}. The SSN-LDA
model assumes that each node is a bag of outgoing links, and models
each outgoing set of links as a mixture over latent components. The
components are the same for each node, but their proportions differ.

As an alternative we introduce a component model for relational data,
where each link is directly assumed to come from a latent component,
and the whole network is a bag of links \citep{Sinkkonen07}. This
model is particularly well suited for modeling of community-type
structure in networks. For conciseness, we call it ICMc (interaction
component model for communities), the latter 'c' reminding of the fact
that it is easy to generate new models from the family of ICMc and
SSN-LDA, with slightly different generative assumptions and
requirements for data.

Both ICMc and SSN-LDA represent a set of links as a probabilistic
mixture over latent components.  Depending on the prior, the models
can find either a given number of latent components, or
nonparametrically adjust the number of components to the data, guided
by a diversity parameter. Moreover, depending on parameters, they are
capable of finding either subnetworks or more graded,
latent-space-like structures.

Both models can be easily and efficiently fitted to data by collapsed
Gibbs sampling \citep{Neal00}, an MCMC technique for sampling from the
posterior where parameters have been integrated out and latent
variables are sampled. In the component models the latent variables
give the assigments of the links to the components. Critical for
successful scaling to large networks is sparseness of representations;
here the component assignments of the links, the variables that are
sampled in the collapsed Gibbs, can be efficiently represented as
sparse arrays, trees, and hash maps.

We compare the two models on two citation networks with a few thousand
nodes, CiteSeer and Cora \citep{SenGetoor07}, and demostrate their
properties on smaller networks.  As a demonstration of a larger-scale
problem, musical tastes of people are derived from the friendship
network of the online music service Last.fm (\texttt{www.last.fm}),
with over 650,000 vertices vertices and almost two million edges.

\begin{figure}[t]
\centerline{
  \begin{minipage}[]{0.16\textwidth}
    \centerline{SSN-LDA}
  \end{minipage}
  \begin{minipage}[]{0.16\textwidth}
    \centerline{ICMc}
  \end{minipage}
  \hspace{0.1\textwidth}
  \begin{minipage}[]{0.25\textwidth}
    \centerline{SSN-LDA}
  \end{minipage}
  \begin{minipage}[]{0.25\textwidth}
    \centerline{ICMc}
  \end{minipage}
}
\centerline{
  \begin{minipage}[]{0.32\textwidth}
    \includegraphics[width=\textwidth]{{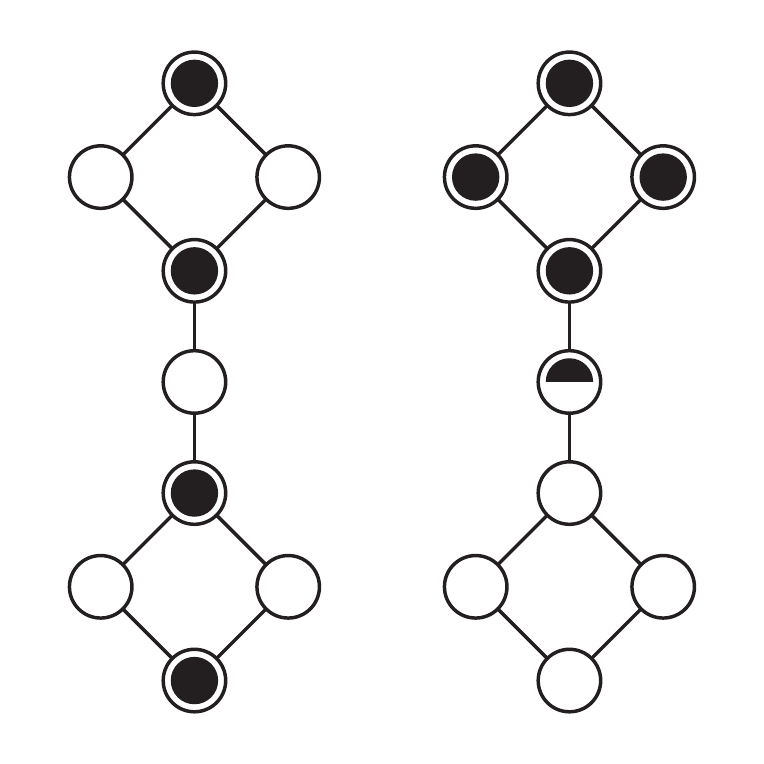}}
  \end{minipage}
  \hspace{0.1\textwidth}
  \begin{minipage}[]{0.25\textwidth}
    \includegraphics[width=\textwidth]{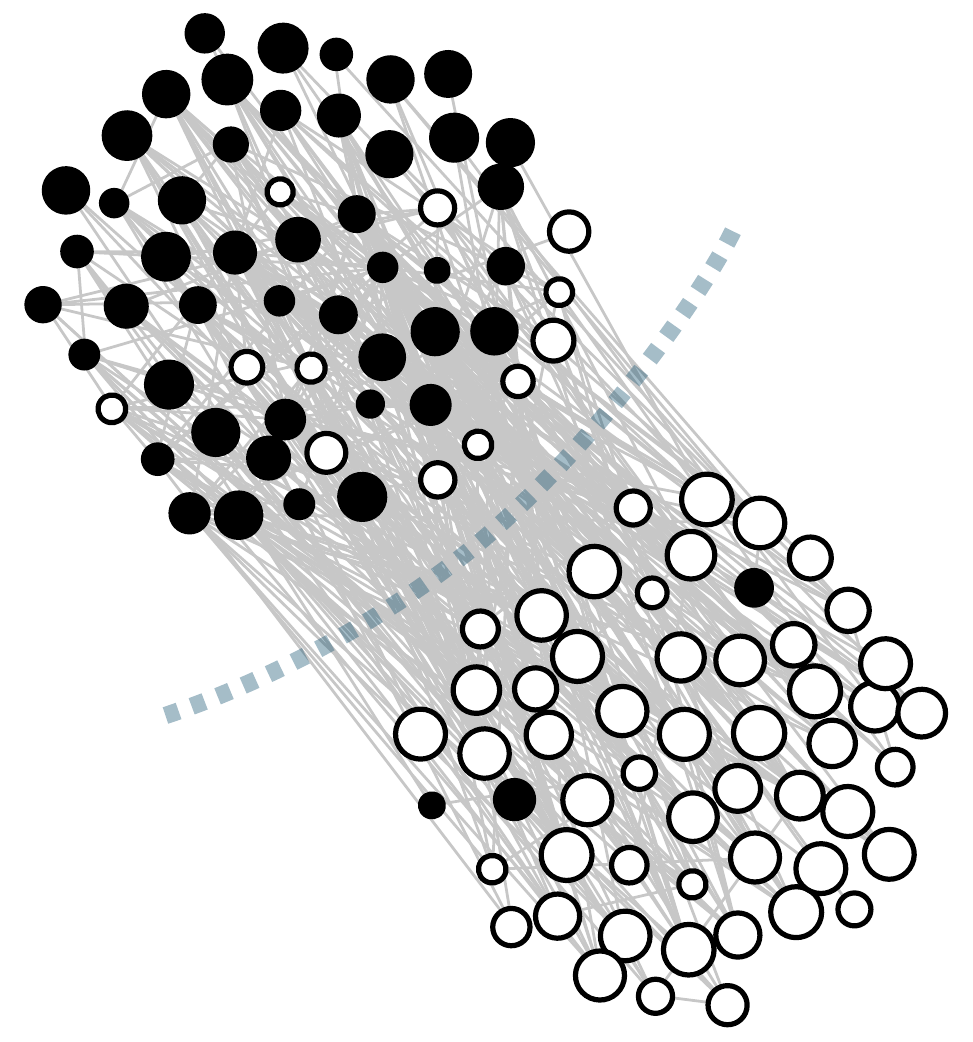}
  \end{minipage}
  \begin{minipage}[]{0.25\textwidth}
    \includegraphics[width=\textwidth]{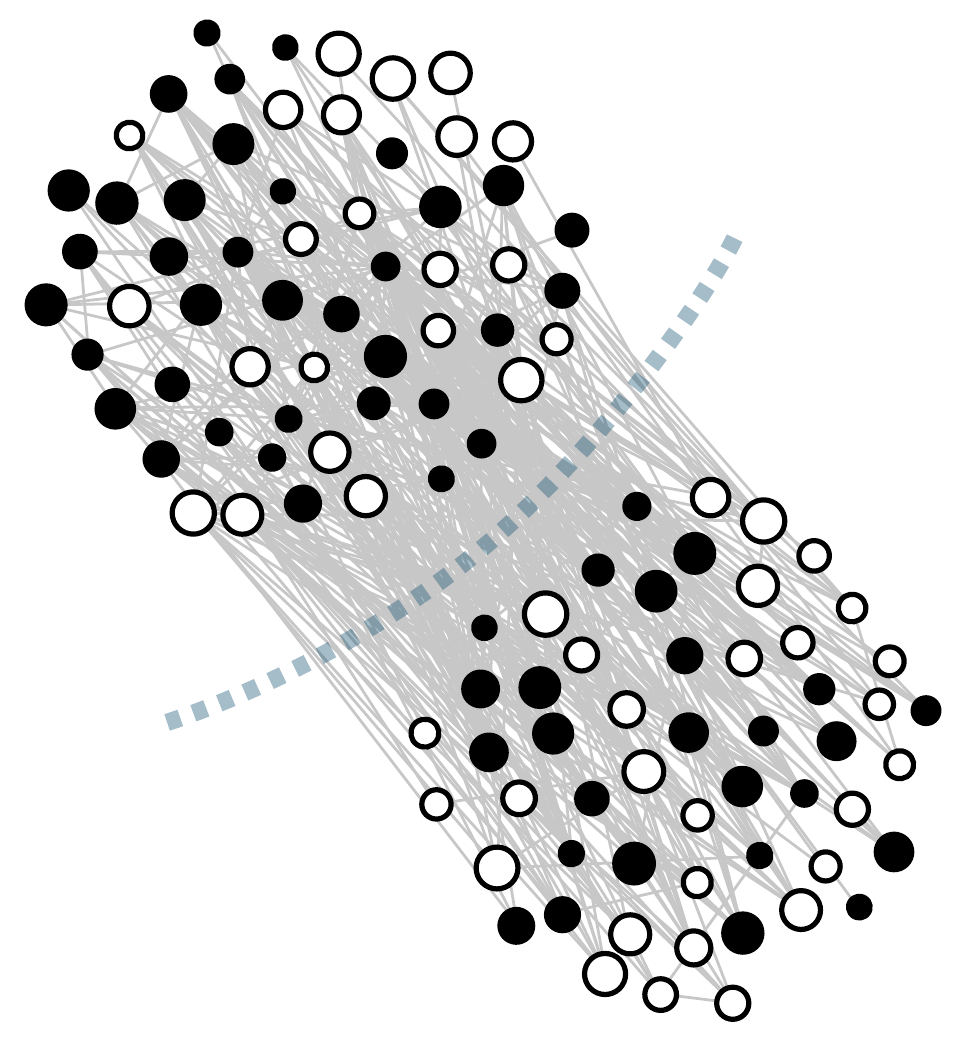}
  \end{minipage}
}
\caption{Network structure and component models. \emph{Left:} A toy
  network split into two components (black and white).  SSN-LDA
  produces both assortative and disassortative components, but here
  favors the disassortative interpretation, grouping nodes by common
  external connectivity. The ICMc solution is assortative and more
  community-like.  \emph{Right:} The Adj-Noun network of adjective and
  noun co-occurrences is bipartite, with a negative modularity of
  -0.241. SSN-LDA, not limited to assortative solutions, finds the
  underlying structure (node size represents certainty). Modularity of
  the SSN-LDA solution (black vs. white) is -0.262 and that of ICMc
  0.188. See Table~\ref{tab:netparams} for details of Adj-Noun.}
\label{fig:disass}
\end{figure}

\section{Two scalable network models}
\label{sec:two}

SSN-LDA models directed links. A unique mixing pattern over latent
\emph{link target profiles} is associated to each node. (Technical
details are presented later, e.g., in Fig.~\ref{fig:plates},
right). The latent profiles correspond to \emph{topics} of text
document models, the original application of LDA. If the node
memberships in latent profiles are sharp enough, that is, if the nodes
are mainly associated to one profile only, the profiles can be
interpreted as subgraphs.  The grouping criterion is a probabilistic
version of the \emph{structural equivalence} principle of sociology
\citep{Michaelson92}: Two nodes belong to the same group if their role
in the network topology is similar, that is, they link to the same
(other) nodes.

In ICMc, a unique mixture over latent components is associated with
each node, and linking is unstructured inside a component. Instead of
structural equivalence, the criterion for subgroups is homogeneous,
symmetric internal connectivity. Link directions are therefore not
modelled. A related social concept is subgroup cohesion
\citep{Wasserman94}, where latent similarity results in connections
inside the group, instead of linking into some common third party. As
a result, the network looks \emph{homophilic} \citep{Lazarsfeld54};
the connected nodes tend to be relatively similar by their non-network
properties.

For technical reasons, the parameterization of linking within a
component in ICMc is in terms of linking probabilities over the
components; memberships of nodes in components can be obtained from
these parameters by the Bayes rule. Equivalently, the model can be
described as modeling the whole graph as a bag of links. Each link
comes from a component specified by a latent variable $z$
(Fig.~\ref{fig:plates}, left). Each component chooses the endpoints of
a link from a component-specific (multinomial) distribution over the
nodes, parameterized by $m_z$.

A further helpful distinction is that of \emph{assortative} and
\emph{disassortative} network properties.  A network is assortative
with respect to a property if the property tends to co-occur in
connected nodes more often than expected by change
\citep{Newman03a}. The opposite, negative correlation in adjacent
nodes, is called disassortativity. SSN-LDA can in principle find
either kinds of structures, while ICMc tends to find only assortative
structure.\footnote{The discussion of the distinction by
  \citet{Newmanleicht06} is indeed applicable to SSN-LDA, for SSN-LDA
  can be seen as a Bayesian extension of the earlier model.}
Modularity, a quality measure for community detection
\citep{Newman06a}, can at least to a degree be used as a measure of
assortativity; if it is negative for a partitioning of the network,
the partitioning is disassortative
\citep{Fortunato07r}. Unfortunately, comparing modularities of
partitionings over different networks is in general not justified
\citep{Fortunato07r}, and hence we cannot use it to compare the
modeling problems.

\begin{figure}
  \begin{minipage}[]{0.49\textwidth}
    \includegraphics[width=\textwidth]{{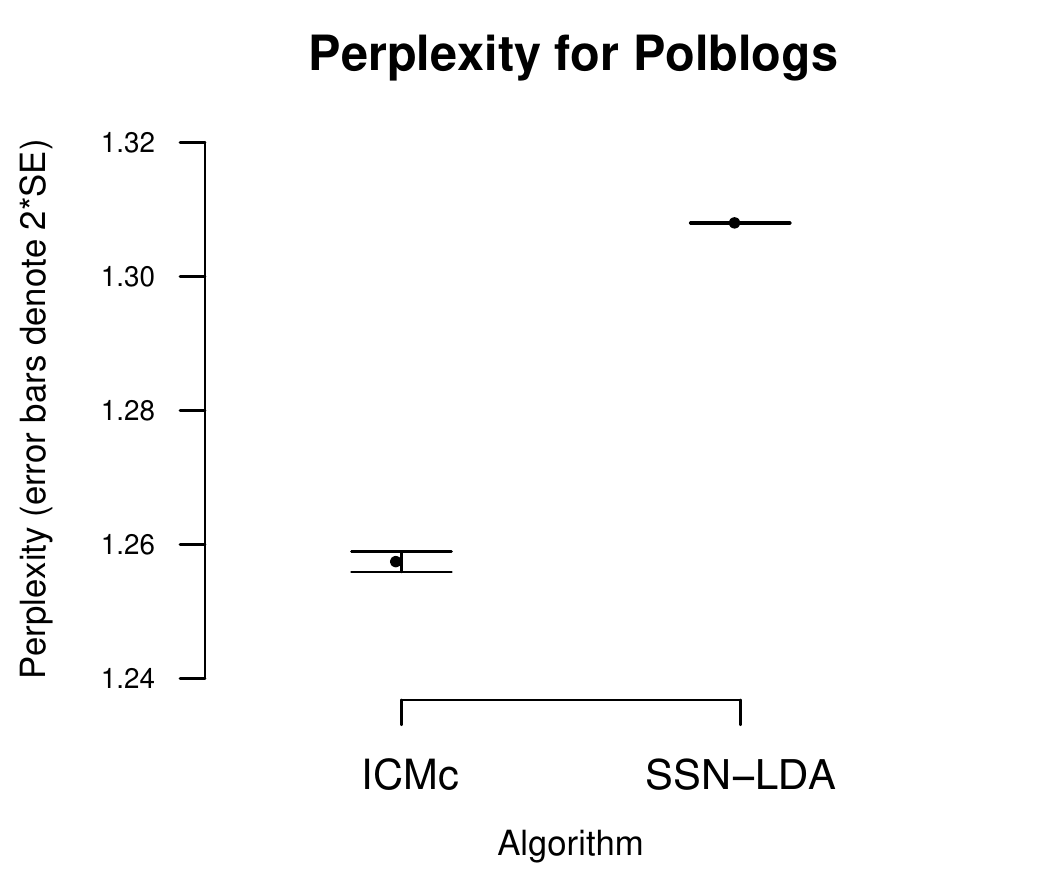}}
  \end{minipage}
  \begin{minipage}[]{0.49\textwidth}
    \includegraphics[width=\textwidth]{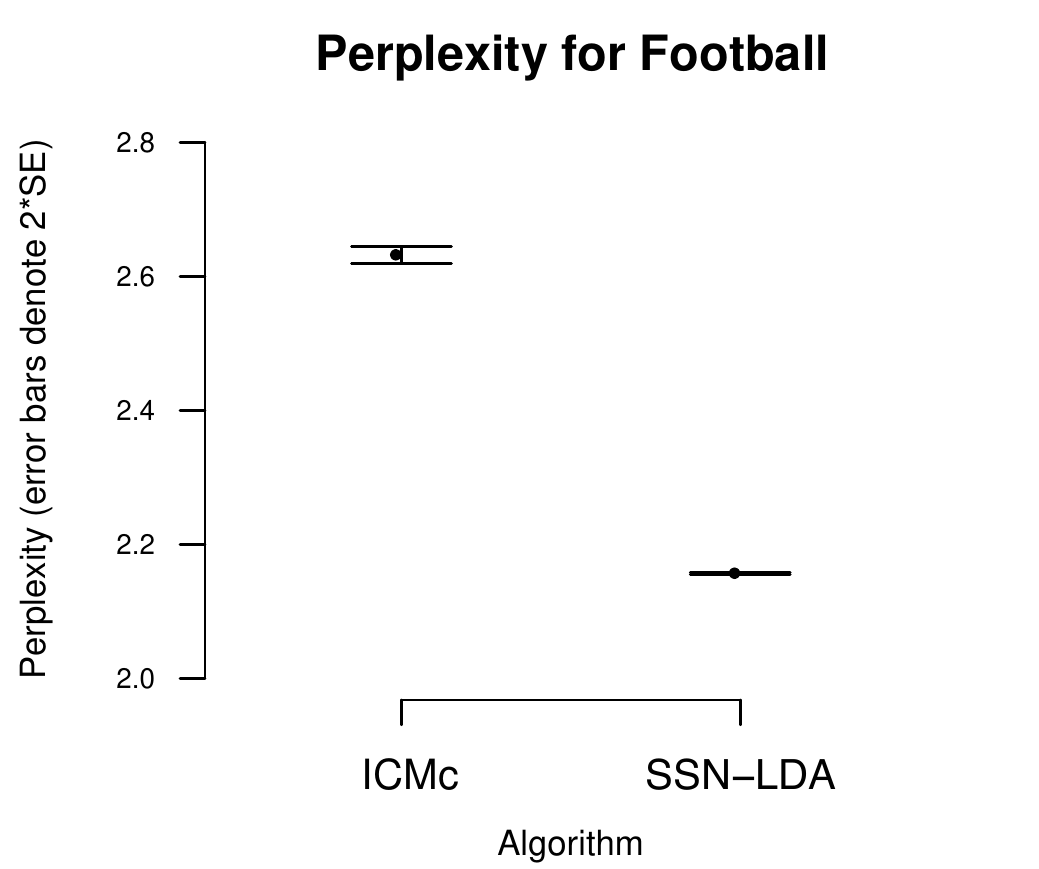}
  \end{minipage}\\
\vspace{0.02\textwidth}\\
\centerline{
  \begin{minipage}[]{0.49\textwidth}
    \centerline{\sf\textbf{ICMc}}
  \end{minipage}
  \begin{minipage}[]{0.49\textwidth}
    \centerline{\sf\textbf{SSN-LDA}}
  \end{minipage}
}
  \begin{minipage}[]{0.49\textwidth}
  \includegraphics[width=\textwidth]{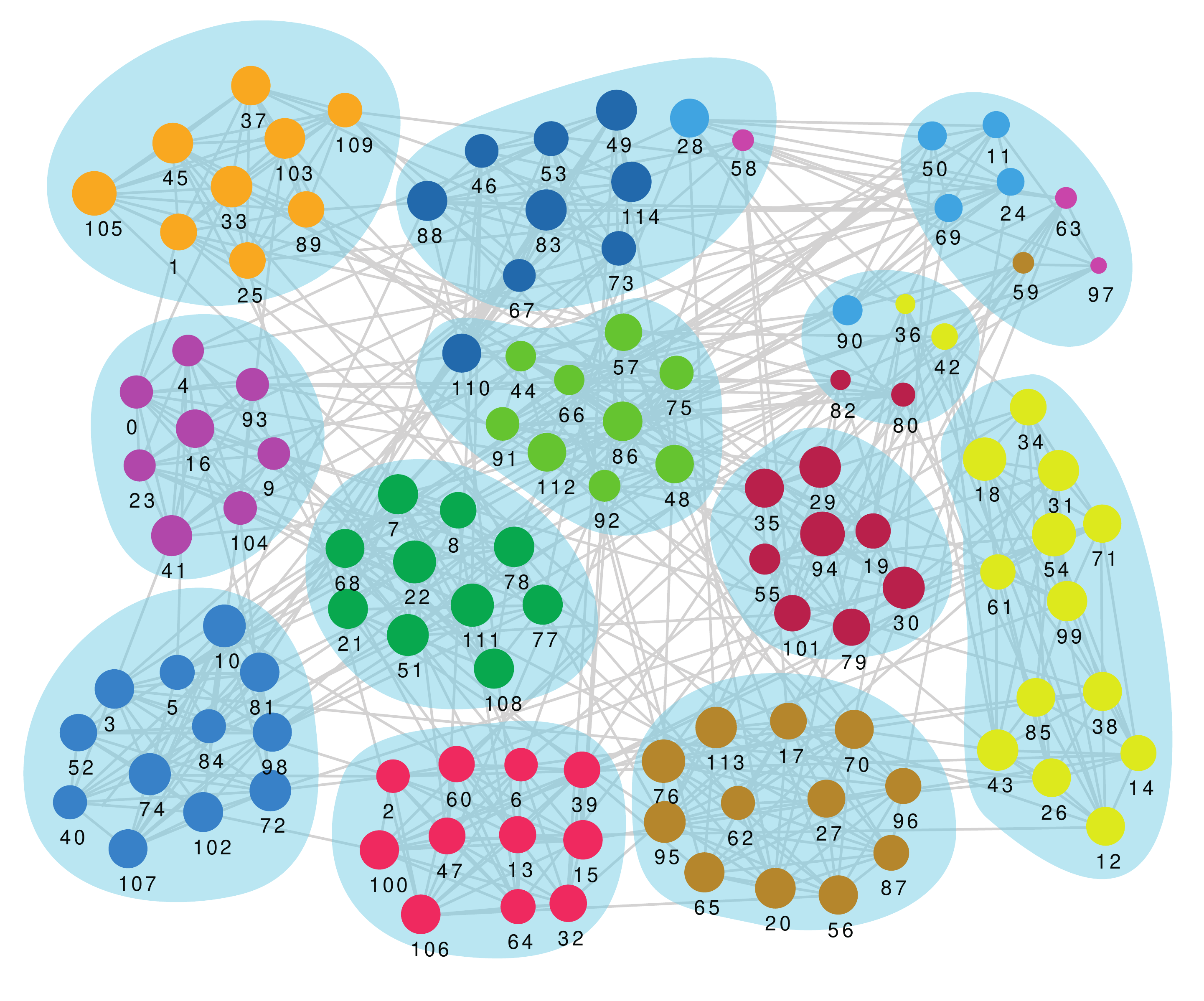}
\end{minipage}
\begin{minipage}[]{0.49\textwidth}
  \includegraphics[width=\textwidth]{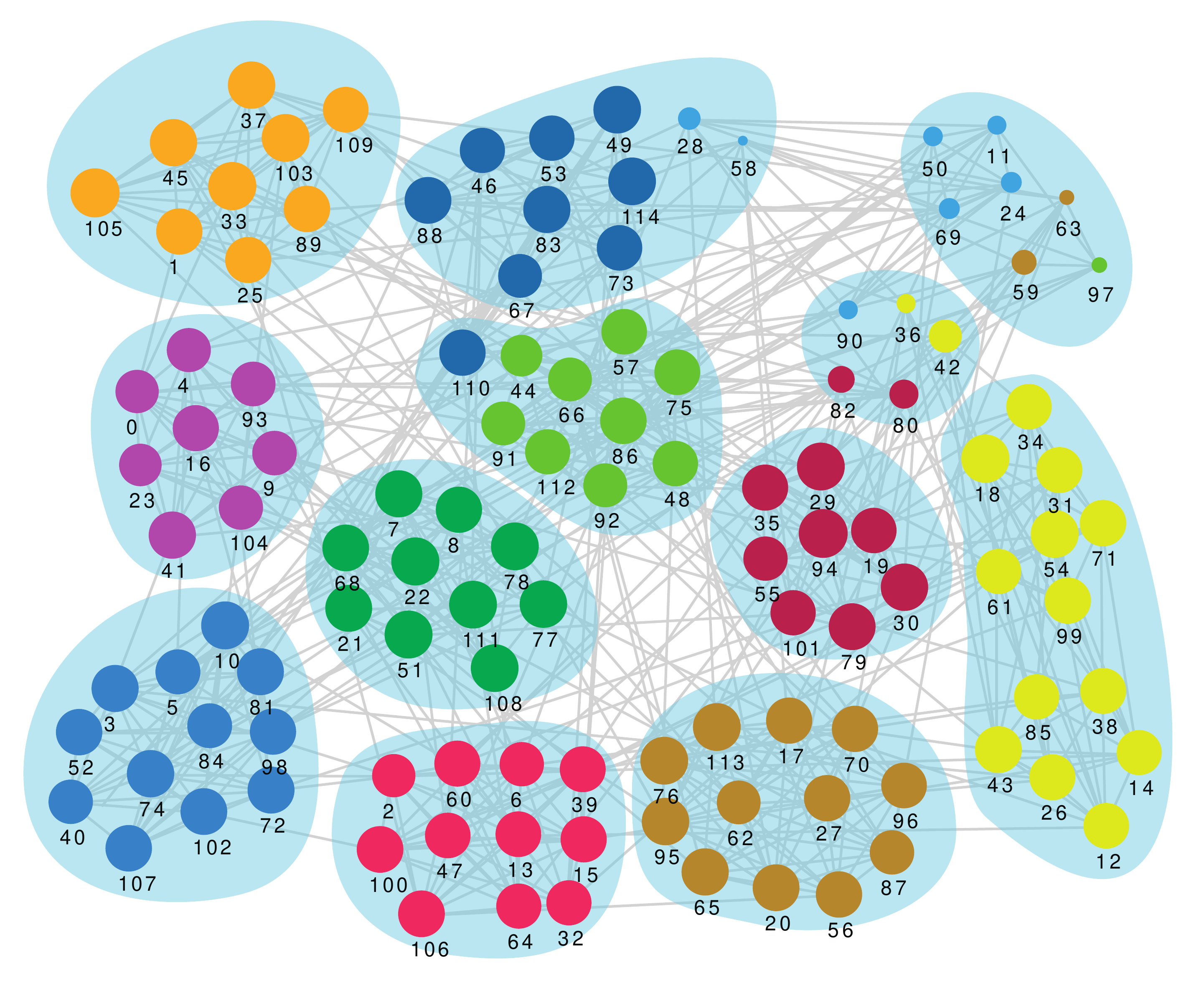}
\end{minipage}
\caption{Relative performance of ICMc and SSN-LDA varies for different
  networks. \emph{Above:} ICMc performs better for a network of
  political blogs, as measured by the perplexity of the components,
  while SSN-LDA is better for a network of US football games. Reasons
  for the differences are unclear, although they might be related to
  the assortativity of the networks with respect to the ground
  clusters (political orientation and football conferences).
  \emph{Below:} Despite perplexity differences, the solutions are
  qualitatively very similar. For the Football network, main
  differences are in the certainty of cluster assignments. Shaded
  areas show the borders of the conferences, while community
  assignments by the model and their certainty are depicted by node
  color and size, respectively. See Table~\ref{tab:netparams} for
  details of the networks and model parameters.}
\label{fig:prob}
\end{figure}

One would expect ICMc to find communities from social and other
networks better than the less specialized SSN-LDA, as long as linking
results from homophily and the communities can be assumed
assortative. The reason is that a model having less degrees of freedom
in its parameterization will be able to more accurately estimate the
parameters from the relatively small observed data sets. On the other
hand, ICMc should be unable to find disassortative structures. This
seems to indeed hold in some extreme cases (Fig.~\ref{fig:disass}),
but in practice differences are often graded and harder to attribute
to properties of the network (Fig.~\ref{fig:prob}).

The behavior of both SSN-LDA and ICMc are determined by their
hyperparameters. Both models can be made to prefer either latent
components of equal size, or to allow heavy size variation. Even more
importantly, either graded or non-overlapping components can be
preferred. In graded components the community membership probabilities
are akin to coordinates in a latent space, while non-overlapping
components divide the nodes sharply into clusters.

Both models can accommodate integer link weights in the sense of
generating multiple links between two nodes.  On the other hand, the
models work particularly efficiently for sparse binary links: If data
is sparse, link probabilties are small overall, multiple links even
more improbable, and the model effectively generates binary data.

The SSN-LDA model is originally based on a finite mixture
\citep{Zhang07isi}, but it is easily extended for a Dirichlet process
prior (DP prior; \citealp{Blackwell73, Neal00}), while ICMc is
originally with a DP prior \citep{Sinkkonen07}, but here applied also
in its finite form.

The models are demonstrated on three small networks in
Figures~\ref{fig:demo} and~\ref{fig:prob}. The first is the Karate
network originating from a study by \citet{Zachary77}. In the study
Zachary observed the social interactions between 34 members of a
karate club over two years. During this period, there was disagreement
among the club members which led to the splitting of the
club. Figure~\ref{fig:demo} demonstrates that ICMc finds the splitting.

The second demonstration network is the Football network
\citep{Girvan02}, which depicts American football games between
Division IA colleges during the fall season 2000. The nodes of the
network represent football teams and edges the games between the
teams. There is a known community structure for the network in the
form of conferences. In general, games between teams that belong to
the same conference are more frequent than games between teams that
belong to different conferences, but sometimes teams prefer to play
mostly against teams in other conferences. Both models find the
structure as seen in Figure~\ref{fig:prob}, SSN-LDA slightly more
accurately. ICMc is somewhat more accurate on another network
derived from political blogs.

%\documentclass[twoside,11pt]{article}
%\usepackage{../jmlr2e}
%\usepackage{amsmath}
%\usepackage{subfigure}
%\begin{document}

\begin{table}
\begin{center}
  \caption[Modeling parameters for smaller networks.]{ Network
    characteristics and modeling parameters for the small and
    medium-size networks.  In the table, \(I\) is the number of nodes
    in the network, \(L\) is the number of edges, \(Q\) the ground
    cluster modularity, and \(\alphadir\) and \(\beta\) are the
    hyperparameters of the models.}
\label{tab:netparams}
\smallskip
\begin{tabular}{lrrr|ll|ll}
&&&& \multicolumn{2}{l|}{ICMc}&\multicolumn{2}{l}{SSN-LDA}\\
Network & \(I\) & \(L\) &\(Q\) &\(\alphadir\)&\(\beta\)&\(\alphadir\)&\(\beta\)\\
\hline
Adj-Noun & 112    & 423    &-0.241& 0.5    & 0.2   &  0.5   & 0.2 \\
Football & 115 & 613 &0.554&0.083&0.03&0.083&0.7\\
Polblogs & 1 222 & 16 714 &0.410&0.5&0.003&0.5&0.4\\
Citeseer & 2 120 & 3 678 &0.517&0.166&0.04&0.166&0.006\\
Cora & 2 485 & 5 067 &0.630&0.143&0.02&0.143&0.025\\
\end{tabular}
\end{center}
\end{table}

%\end{document} 

\subsection{Interaction Component Model for Communities (ICMc)}

\begin{figure}
  \begin{minipage}[]{0.49\textwidth}
  \includegraphics[width=\textwidth]{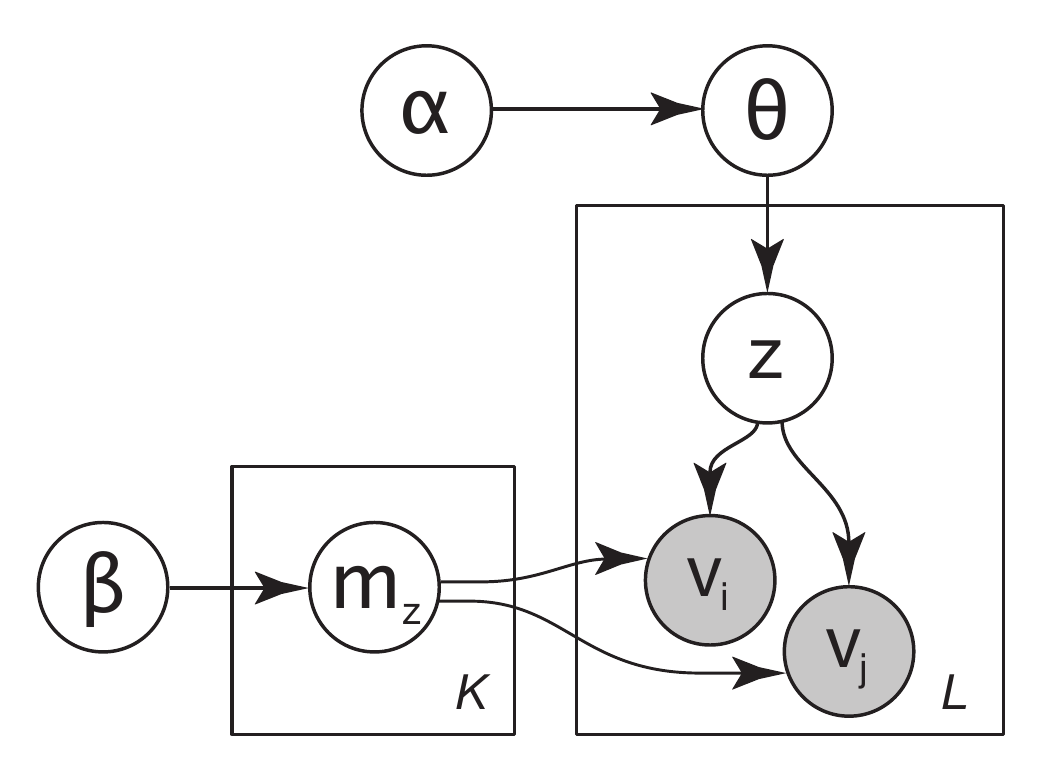}
\end{minipage}
\begin{minipage}[]{0.49\textwidth}
  \includegraphics[width=\textwidth]{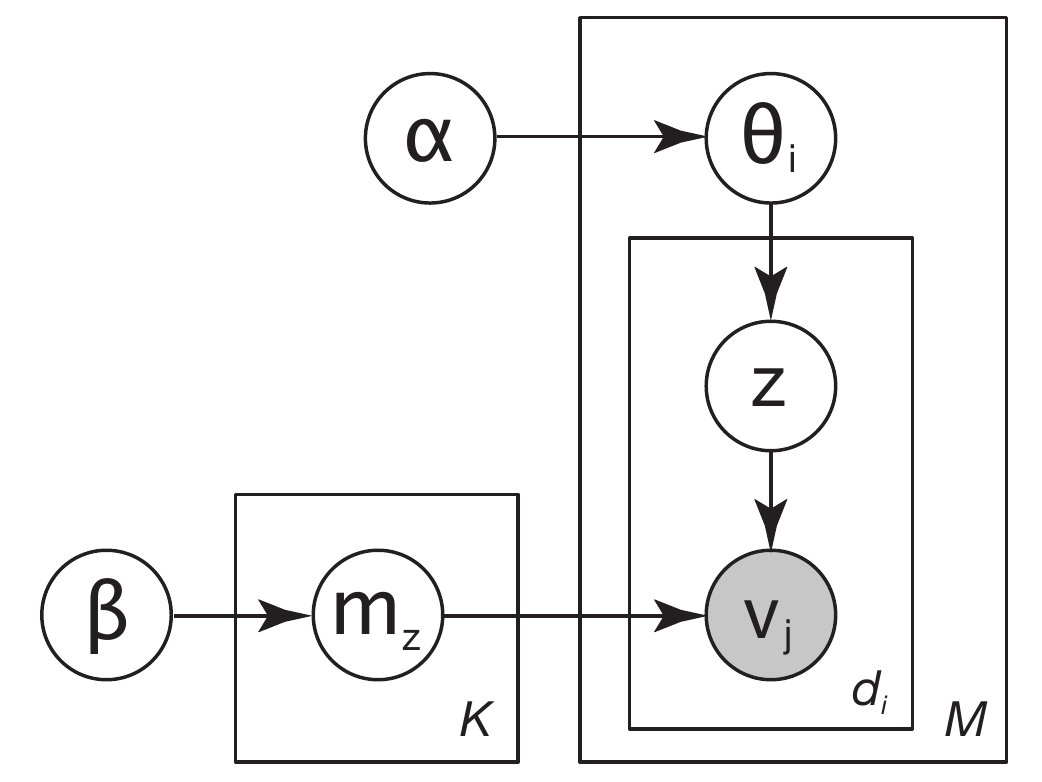}
\end{minipage}
\caption{The ICMc model (\emph{left}) and SSN-LDA
  (\emph{right}). SSN-LDA is effectively the Latent Dirichlet
  Allocation (LDA; \citealp{Buntine02,Blei03}) applied to network
  data, with nodes playing both the role of 'words', at the receiving
  end of links, and 'documents' at the sending end. ICMc has no
  hierarchy level for nodes. Instead, it generates \emph{two} nodes
  for each link; the links are undirected. See Section \ref{sec:two}
  for the notation and further discussion.}
\label{fig:plates}
\end{figure}

The generative process out of which the network is supposed to arise
is the following (see Fig.~\ref{fig:plates} for a diagram); it is
parameterized by the hyperparameters $(\alpha,\beta)$.
\begin{itemize}
\item[(1.1)] Generate a multinomial distribution $\theta$ over
  \emph{latent components} $z$. For $K$ components, the multinomial is
  generated from a $K$-dimensional Dirichlet distribution with
  all parameters set to $\alphadir$, $\Dir_\mathrm{sym}^K(\alphadir)$, 
  and for an infinite number of components from the Dirichlet process
  $\DP(\alphadp)$.
\item[(1.2)] To each $z$, associate a multinomial distribution over the
  $M$ vertices $i$ by sampling the multinomial parameters $m_{z}$
  from the Dirichlet distribution $\Dir_\mathrm{sym}^M(\beta)$.  (To
  clarify, we have $\sum_i m_{zi} = 1$ for each $z$, and $\sum_z
  \theta_z=1\,$.); 
\item[(2)] Then repeat for each link $l=1\dots L$: 
\begin{itemize}
\item[(2.1)] Draw a latent component $z$ from the multinomial
  $\theta$.
\item[(2.2)] Choose two nodes, $i$ and $j$, independently of each
  other, with probabilities $m_z$; set up a nondirectional link between
  $i$ and $j$.
\end{itemize}
\end{itemize}

Within components, edges are generated independently of each other;
the non-random structure of the network emerges from the tendency of
components to prefer certain vertices (that is, $m$). In contrast to
many other network models, the latent variables operate on the edge
level, not on the vertex level. There is no explicit hierarchy level
for vertices, but because vertices typically have several edges, they
are implicitly treated as mixtures over the latent
components. Finally, the model is parameterized to generate self-links
and multi-edges because this choice allows sparse implementations
which would not be directly possible with a potential alternative
model that would generate binary links from the Bernoulli
distribution.

Although in the case of a Dirichlet process prior the number of
potentially generated components is infinite, the prior gives an
uneven distribution over the components. Therefore, with a suitably
small value of $\alphadp$, we observe much fewer components than the
number of links is, and the model is useful. On the other hand,
$\beta$ describes the unevenness of the degree distribution of the
nodes \emph{within components\/}: a high $\beta$ tends to give
components spanning over all nodes, while a small $\beta$ prefers
mutually exclusive, community-like components.

We have estimated the model with Gibbs sampling \citep{Geman84}, a
variant of MCMC methods that produce samples from the posterior
distribution of the model parameters and the latent component
memberships. As a side note, maximum likelihood or MAP estimation of
the model is not sensible since the number of parameters and latent
variables is large compared with the available data, It is easy to
derive an EM algorithm for the finite-mixture ICMc, but it gets stuck
into suboptimal local posterior maximums at the borders of the
parameter space.

We use Gibbs sampling with some of the model parameters integrated
out, called Rao-Blackwellized, or collapsed \citep{Neal00}. (For the
joint distribution of the model and the derivation of the estimation
algorithm see the Appendix). In the collapsed Gibbs estimation
algorithm the unknown model parameters $m_{zi}$ and $\theta_z$ are
marginalized away and only the latent classes of the edges, $z_l$, are
sampled, one edge at a time.  In general, we denote edge counts per
component by $n_z$, component-wise vertex degrees by $k_{zi}$, and the
endpoints of the left-out edge by $(i, j)$. Then delete one edge,
resulting in counts $(n', k', N')$ that are equal to $(n, k, N)$ but
without the one edge. The component probabilities of the left-out edge
are
\begin{equation}
p(z|i, j) \propto 
\frac{k'_{zi}+\beta}{2n'_z+1+M\beta} 
\times 
\frac{k'_{zj}+\beta}{2n'_z+M\beta} 
\times 
\frac{n'_z+\alphadir}{N'+K\alphadir}
\label{eq:urn}
\end{equation}
for the Dirichlet prior, and
\begin{equation}
p(z|i, j) \propto 
\frac{k'_{zi}+\beta}{2n'_z+1+M\beta} 
\times 
\frac{k'_{zj}+\beta}{2n'_z+M\beta} 
\times 
\frac{\C(n'_z,\alphadp)}{N'+\alphadp}
\label{eq:urndp}
\end{equation}
for the Dirichlet process prior. The 'chooser' function $\C(n_z,
\alphadp)\equiv n_z$ if $n_z \neq 0$ and $\C(0, \alphadp)=\alphadp$.
The case with $\alphadp$, as opposed to $n_z$, corresponds to a new
component with no other links so far.

This sampling step is simply repeated iteratively for all links, until
convergence to the posterior distribution, or until the results are
satisfactory by some other measure.  A particularly elegant, although
not necessarily the most efficient initialization of the sampler
starts from empty urns, with $k_{zi}=n_z=N=0$, then runs through the
edges once in a random order and populates the urns according to
(\ref{eq:urn}) or (\ref{eq:urndp}) while counting only the edges seen
so far.

\begin{algorithm}[tb]
\caption{$\proc{ICMc-Gibbs}$. \; A simple implementation for the ICMc
  algoritm with a Dirichlet prior.} 
\label{alg:simpleimplementation}

%\vspace{0.05in}

\begin{codebox} 
\Procname{$\proc{Simple-Gibbs-Sampling}(\alphadir, \beta, L)$}
\zi $t_{nodes} \gets $ node count 
\zi \For{$c \gets 1$ \To \id{components}} \Do
\RComment initialize data structures
\zi $A[c] \gets 0$
\zi \For{$n \gets 1$ \To $t_{nodes}$} \Do $\; K[n, c] \gets 0$
\End
\End
\zi \For{$i \gets 1$ \To \id{iterations}} \Do
\RComment main iteration loop
\zi \ForEach{$l$ \In $L$} \Do
\zi $v_i, v_j \gets $ first and second node of $l$
\zi \If{${i \neq 1}$} \Do 
\zi $z_{old} \gets Z[l]$
\zi decrement $K[v_i, z_{old}], K[v_j, z_{old}], A[z_{old}]$
\End
\zi \For{$c \gets 1$ \To \id{components}} \Do
\zi $P[c] \gets \proc{Calc-Probability}(A[c], K[v_i, c], K[v_j, c], t_{nodes}, \alphadir, \beta)$
\End
\zi $z_{new} \gets$ sample index from $P$
\zi $Z[l] \gets z_{new}$
\zi increment $K[v_i, z_{new}], K[v_j, z_{new}], A[z_{new}]$
\End
\End 
\zi \Return K, Z
\end{codebox}

%\vspace{0.05in}

\begin{codebox} 
\Procname{$\proc{Calc-Probability}(n_c, k_a, k_b,t_{nodes}, \alphadir, \beta)$} 
\zi \Return $\dfrac{(k_a+\beta) (k_b+\beta) (n_c+\alphadir)}{
  (2n_c+1+\beta  t_{nodes}) (2n_c+\beta  t_{nodes})}$
\end{codebox} 

\end{algorithm} 

The goal of model fitting is usually to infer community memberships of
the nodes. From the Bayes rule we obtain
\begin{equation}
p(z|i) = \frac{\theta_z m_{zi}}{\sum_{z'} \theta_{z'} m_{z'i}}\;.
\label{eq:membayes}
\end{equation}
A sample of the marginalized parameters $\theta$ and $m$ can be
reconstructed from each realization of the counts $(k, n)$ by sampling
from the conditional Dirichlet distributions given the priors and the
counts:
\begin{equation}
\theta \sim \Dir_z(n_z + \alphadir) 
\,\,\,\mathrm{or } \,\,\, 
\theta \sim \Dir_z(\left\{n_z,\alphadp\right\}_z)\,,
\,\,\,\mathrm{and } \,\,\, 
m_z \sim \Dir_i(k_{zi}+\beta)
\;.
\end{equation}
Note that $n_z=\sum_{i} k_{zi}/2$. In the case of the Dirichlet
process prior, the parameter $\theta$ has probabilities of the
components with at least one assigned link, and then the probability
of all empty components summed up into the last bin. These correspond
to the Dirichlet parameters $n$ and $\alphadp$, respectively.

Even if one wants to reconstruct $\theta$ and $m$, collapsed Gibbs is
likely to be faster than full Gibbs. The reasons are twofold: Firstly,
Gibbs converges faster when the parameter updates are not in the main
loop. Secondly, one usually uses decimation in sampling from the
converged chain, and the $(\theta, m)$ need to be constructed only for
the decimated samples.

It is often sufficient to estimate the community memberships from
the expected values of the marginalized parameters,
\begin{equation}
\hat \theta = \frac{n_z+\alphadir}{\sum_{z'} n_{z'} + K\alphadir}
\,\,\,\mathrm{or } \,\,\, 
\hat \theta = \frac{n_z}{\sum_{z'} n_{z'} + \alphadp}
\;,
\end{equation}
and
\begin{equation}
\hat m_z = \frac{k_{zi}+\beta}{\sum_{i'} k_{zi'} + M\beta}
\;.
\end{equation}
Substituting the expectations into (\ref{eq:membayes}), we find that
for small $\alpha$ and $\beta$,
\begin{equation}
p(z|i) \approx \frac{k_{zi}}{\sum_{z'} k_{z'i}}
\label{eq:icm-nodemem}
\end{equation}
is a good approximation.

Prediction for new data is straightforward; the component memberships
of the links associated to a new node can be sampled from
(\ref{eq:urn}), given old links. If the new links are not conditionally
independent given the old data, one can run a short Gibbs iteration on
the new links. 

\subsection{SSN-LDA}

SSN-LDA \citep{Zhang07isi} also has two hyperparameters, denoted by
$\alpha$ and $\beta$, but they are in a slightly different role than
in ICMc (see Fig.~\ref{fig:plates}). The generative process is as
follows:
\begin{itemize}
\item[(1.1)] Generate $M$ multinomial distributions $\theta_{i}$,
  $i=1,\dots, M$, over \emph{latent components} $z$, $z=1,\dots, K$,
  either from a $K$-dimensional Dirichlet distribution
  $\Dir_\mathrm{sym}^K(\alphadir)$, or from the Dirichlet process
  $\DP(\alphadp)$.
\item[(1.2)] Assign a multinomial distribution $m_{z}$ over the
  vertices $i$ to each component $z$ by sampling from the Dirichlet
  distribution $\Dir_\mathrm{sym}^M(\beta)$.
\item[(2)] Then repeat for each link $l=1,\dots, L$, with sending nodes
  $i=1,\dots, M$: 
\begin{itemize}
\item[(2.1)] Draw a latent component $z$ from the multinomial
  $\theta_{i}$.
\item[(2.2)] Choose the link endpoint $j$ with probabilities $m_z$;
  set up a directional link between $i$ and $j$.
\end{itemize}
\end{itemize}
We have presented the generative process of links in a
flat form to make comparison to ICMc easier; in step 2, the loop over
nodes is avoided by referring to the node indices $i$ associated to
links.

In contrast to ICMc, SSN-LDA has the node as an explicit hierarchy
level---in the generative model, there are the parameters $\theta$ for
each node separately, and $\alpha$ is the common hyperparameter of
these node-wise distributions. As in ICMc, the hyperparameters $m$ are
associated to latent components over nodes, and $\beta$ determines
their prior. But now $m$ determines only the probabilities of the
\emph{receiving nodes}. Sending probabilities, associated to starting
points $i$ of the links, are modeled by $\theta$.

Collapsed Gibbs sampling operates on two sets of counts: $n_{iz}$ that
counts the sender--component combinations $(i, z)$ for links,
originating from step (2.1) of the generative process, and $k_{zj}$
counting the receiver--component combinations $(j,z)$ from step (2.2)
of the process.  Following \citet{Griffiths04}, and
\citet{Zhang07isi}, the conditional probabilities for sampling a
left-out link in a collapsed Gibbs iteration, given hyperparameters
and all other links, is
\begin{equation}
p(z|i, j) \propto 
\frac{k'_{zj}+\beta}{k'_{z\cdot}+M\beta} 
\times 
\frac{n'_{iz}+\alphadir}{n'_{i\cdot}+K\alphadir}
\;,
\label{eq:ldaurn}
\end{equation}
where sums over counts $k'$ and $n'$ have been denoted by the dot
notation.  We have omitted the derivation of the Dirichlet process
variant, because it is very similar to the derivation of the DP ICMc
(see the Appendix), leading to:
\begin{equation}
p(z|i, j) \propto 
\frac{k'_{zj}+\beta}{k'_{z\cdot}+M\beta} 
\times 
\frac{\C(n'_{iz},\alphadp)}{n'_{i\cdot}}
\;.
\label{eq:ldaurndp}
\end{equation}
Again, parameter reconstruction for $\theta$ and $m$ can be done
either by sampling from the corresponding Dirichlet distributions, or
by computing the conditional MAP estimates, either roughly or exactly
including priors. As with ICMc, we have used the rough alternative
suitable for small values of $\alpha$ and $\beta$:
\begin{equation}
p(z|i) \approx \frac{n_{iz}}{n_{i\cdot}}\;.
\label{eq:lda-nodemem}
\end{equation}
This is for the community memberships of nodes as senders of
links. Because in SSN-LDA links are directed, it is possible to define the
memberships also in terms of received links,
\begin{equation}
p(z|j) \approx \frac{k_{zj}}{k_{\cdot j}}\;.
\end{equation}

%directed-force layout with the \emph{Prefuse\/} software package.

\subsection{Efficient implementation of the collapsed Gibbs samplers}
\label{sec:alg}

Large real-life networks are sparse almost by definition, and for
efficiency it is important to preserve the sparseness in model
structures. ICMc and SSN-LDA facilitate sparse structures, since
likelihoods decompose into sums over existing links, and terms related
to non-links do not appear.

Collapsed Gibbs sampling of ICMc and SSN-LDA needs tables for $n$ and $k$
which together, as a first approximation, are of the size complexity
$O(MK)$. In addition one needs to keep track of the component
identities of the links, an array of size $O(L)$.  But in both models
the degree of a node poses an upper limit for its component
heterogeneity, so that only a few of the counts $k$, or $k$ and $n$ in
LDA, are simultaneously non-zero, allowing sparse representation of
the count tables.  Therefore with hash tables memory consumption can
be reduced to $O(M\bar d+L+K)$ where $\bar d$ is the average degree of
a node. Because $\bar d=L/M$, memory consumption scales as $O(L+K)$.

Marginal sums of the count tables, notably $n$ in ICMc, can be
represented in a sparse form and updated efficiently during sampling
with the aid of a self-balancing binary tree. The idea of using a tree
in sampling of discrete distributions was originally proposed by
\citet{Wong80}, and another method for using binary trees in
simulations is provided by \citet{Blue95}. In our implementation the
Arne-Andersson tree (AA tree) is used \citep[see, e.g.,][]{Weiss98}, but
other self-balancing binary trees would be equivalent in
performance. A partial sum tree is formed, where in each node, the
total probability of the node is stored, together with the sums of
probabilities of both the left and right children of the node. When the
probability of a node is changed, the modifications are propagated up
to the parents of the node. Sampling proceeds recursively down the
tree as a sequence of weighted Bernoulli samples. 

These sparse representations, and the binary tree for the marginal
sums, make it possible to run models with at least tens of thousands
of components in an ordinary PC or server. These structures also fit
well with the dynamic component numbers due to the Dirichlet process
prior. With the data structures described above, running time per one
Gibbs iteration over all the nodes becomes $O(L\,\bar d\; \log
K)$. That is, the time needed for an iteration scales linearly in the
number of edges and logarithmically in the number of components.

% Hand waving
It is hard to give any general rule on the number of Gibbs iterations
needed for convergence. Because the variables in the collapsed Gibbs
algorithm correspond to links, the dependency graph of the variables
is like the original network, but with the nodes being in the role of
links, and vice versa. The path lenghts of the dependency network are
therefore proportional to the path lengths of the original
network. Let us assume that the average path length scales as $l \propto
\log M$, as is the case with many small-world networks
\citep{Albert02}. In Gibbs, information diffusion over the network can
be expected to take $l^2$ iterations, analogously to ordinary
diffusion. This leads to the conjecture that the number of Gibbs
iterations should be proportional to $\log^2 M$.

\section{Tests}
\label{sec:t}

We compared SSN-LDA and ICMc on two medium-scale social network
datasets, Cora and Citeseer \citep{SenGetoor07}, in the task of
finding a predefined set of known clusters.  Performance on large
networks of $10^5 \dots 10^6$ nodes is then demonstrated for one of
the models (ICMc) with two friendship networks from the music site
Last.fm.

\subsection{ICMc vs. SSN-LDA}
\label{sec:comp}

The Cora and the CiteSeer datasets consist of content descriptions of
scientific publications and citations between them. The Cora dataset
has 2,708 papers in seven predefined classes, while the CiteSeer
dataset contains 3,312 publications in six classes. We used only
the citation information, and the predefined classes as a ground truth
for clustering. Nodes (publications) not belonging to the main
components of the network were removed, and directional links were
symmetrized. The resulting network for Cora has 2,120 nodes and for
Citeseer 2,485 nodes (Table~\ref{tab:netparams}).

\begin{figure}
  \begin{minipage}[]{0.57\textwidth}
  \includegraphics[width=\textwidth]{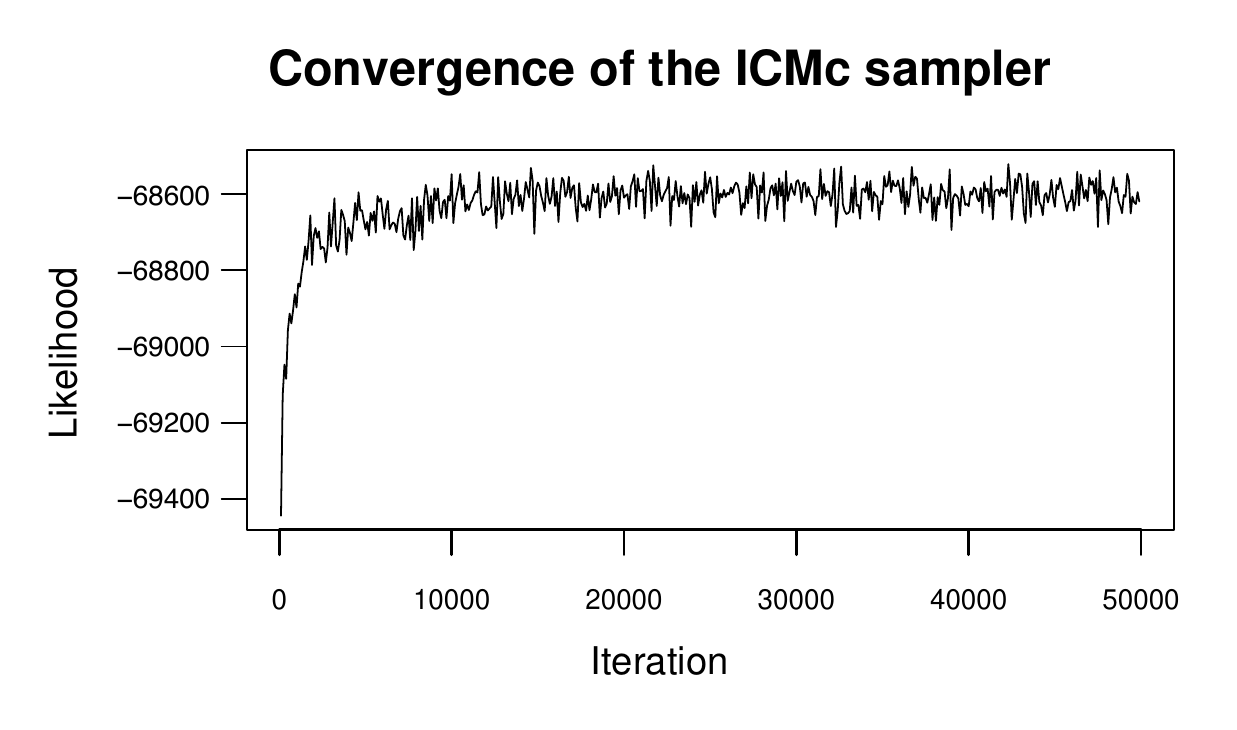}
\end{minipage}
\begin{minipage}[]{0.42\textwidth}
  \includegraphics[width=\textwidth]{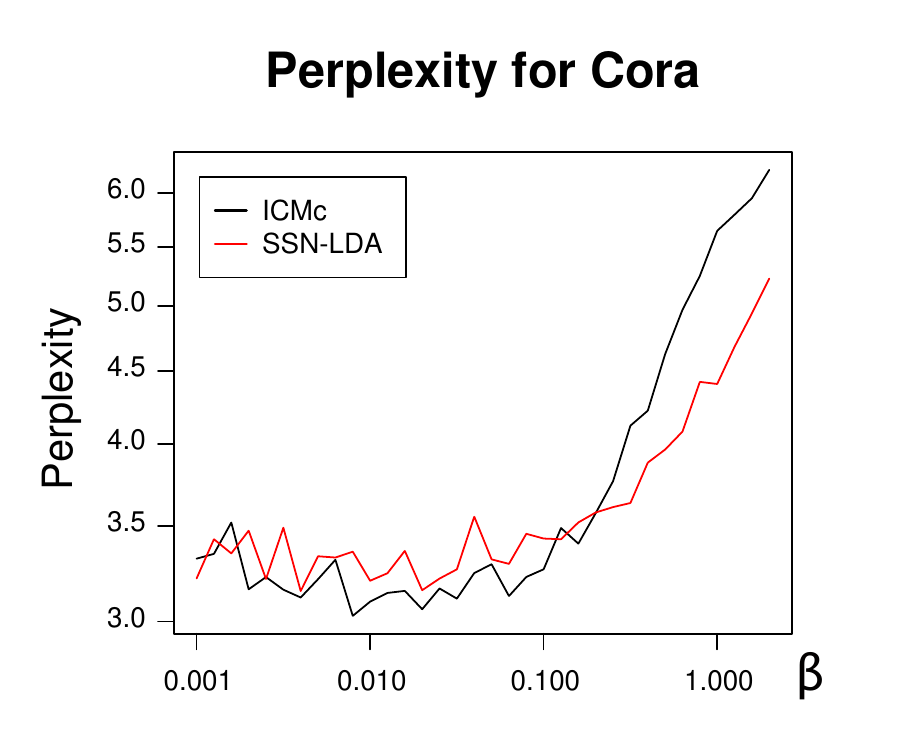}
\end{minipage}
\caption{Gibbs samplers on the Cora citation set: convergence and
  sensitivity to the hyperparameter $\beta$. \emph{Left:\/}
  Leave-one-out logarithmic posterior probability of the data for a
  single ICMc chain. This can be recorded easily during sampling as
  log probabilities of the drawn link assignments.  We ran 50,000
  iterations over the data, but about 15,000 would have been
  enough for convergence, and about 3,000 for getting useful
  results. SSN-LDA convergence was very similar. \emph{Right:\/}
  Perplexity for the Cora dataset with a range of hyperparameter
  values \(\beta\). Each reported value is an average of four
  chains. Both models are quite robust with respect to $\beta$. }
\label{fig:whatever}
\end{figure}

Following \citet{Zhang07isi} and our own experiences
\citep{Aukia07}, we fixed $\alphadir=1/K$ for both models and
datasets. Values for the parameter $\beta$ were chosen with pretests
(Table~\ref{tab:netparams} and Fig.~\ref{fig:whatever}). In general,
the models with a Dirichlet prior and a small number of components are
quite insensitive to values of $\alphadir$ and $\beta$ within the range
$0.001\dots 0.1$.

The Gibbs sampler was initialized as suggested in
Section~\ref{sec:alg} and run for 50,000 iterations (see
Fig.~\ref{fig:whatever}). We then took 100 samples at intervals of
100. Each sample consists of the latent cluster memberships $z$ for
all links. Node memberships were constructed by (\ref{eq:icm-nodemem})
and (\ref{eq:lda-nodemem}) for each sample separately, and these were
summed up to get confusion matrices. 

\begin{figure}[t]
  \begin{minipage}[]{\textwidth}
  \includegraphics[width=.49\textwidth]{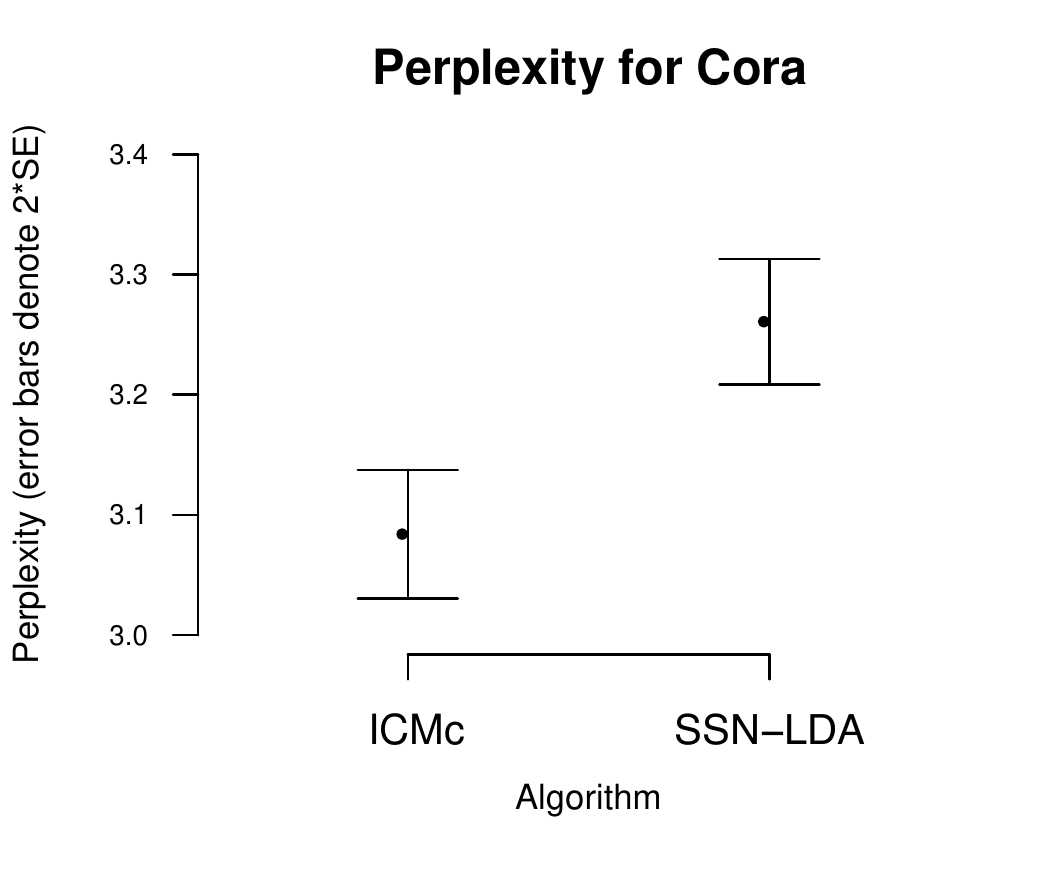}
  \includegraphics[width=.49\textwidth]{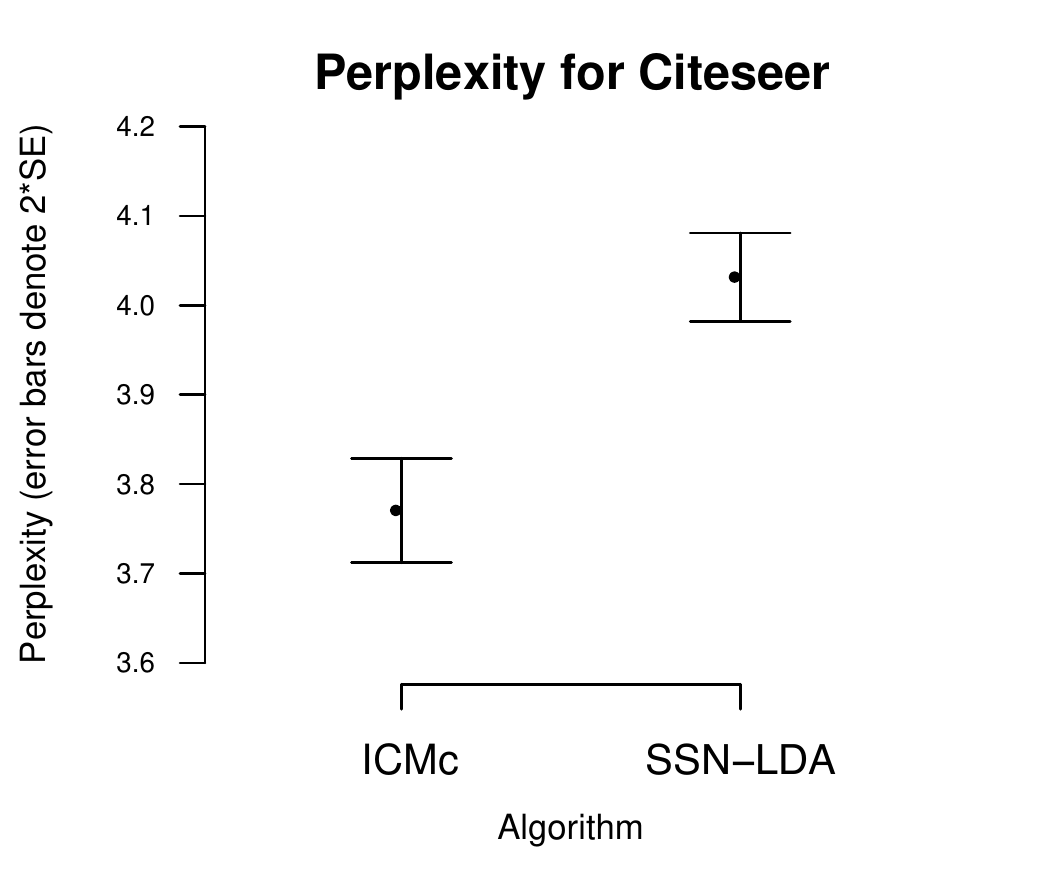}
\end{minipage}
\begin{minipage}[]{0.49\textwidth}  
  \includegraphics[width=.49\textwidth]{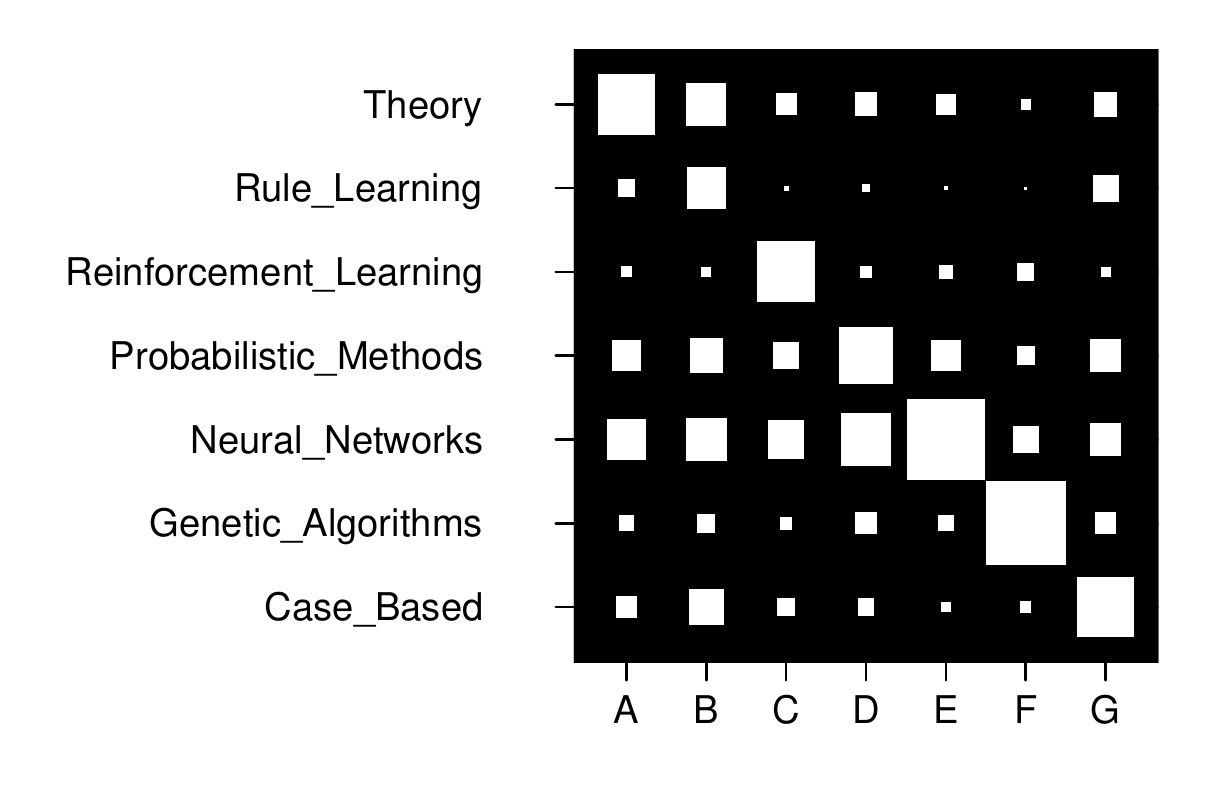}
  \includegraphics[width=.49\textwidth]{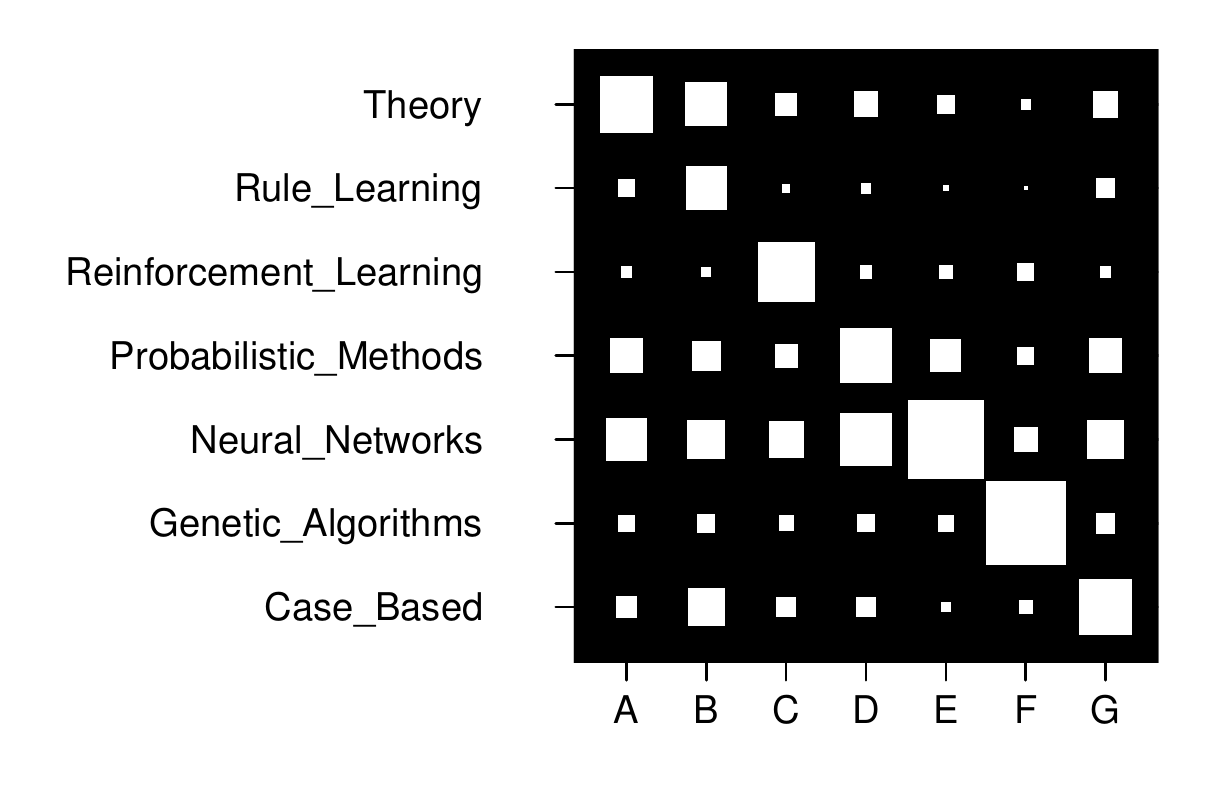}
\end{minipage}
\begin{minipage}[]{0.49\textwidth}
  \includegraphics[width=.49\textwidth]{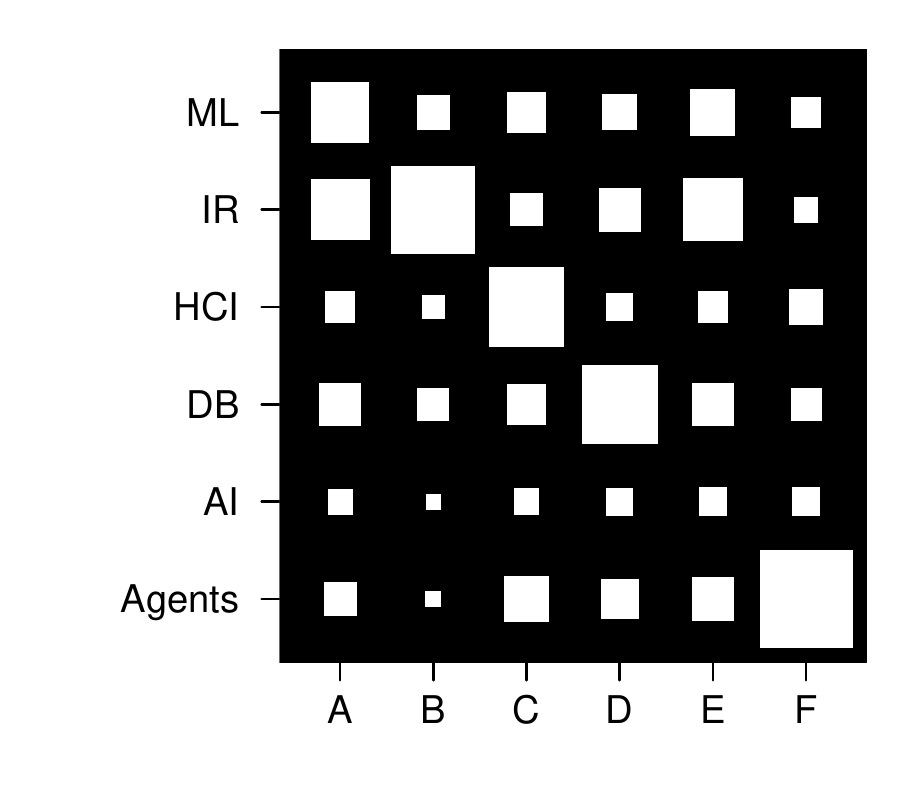}
  \includegraphics[width=.49\textwidth]{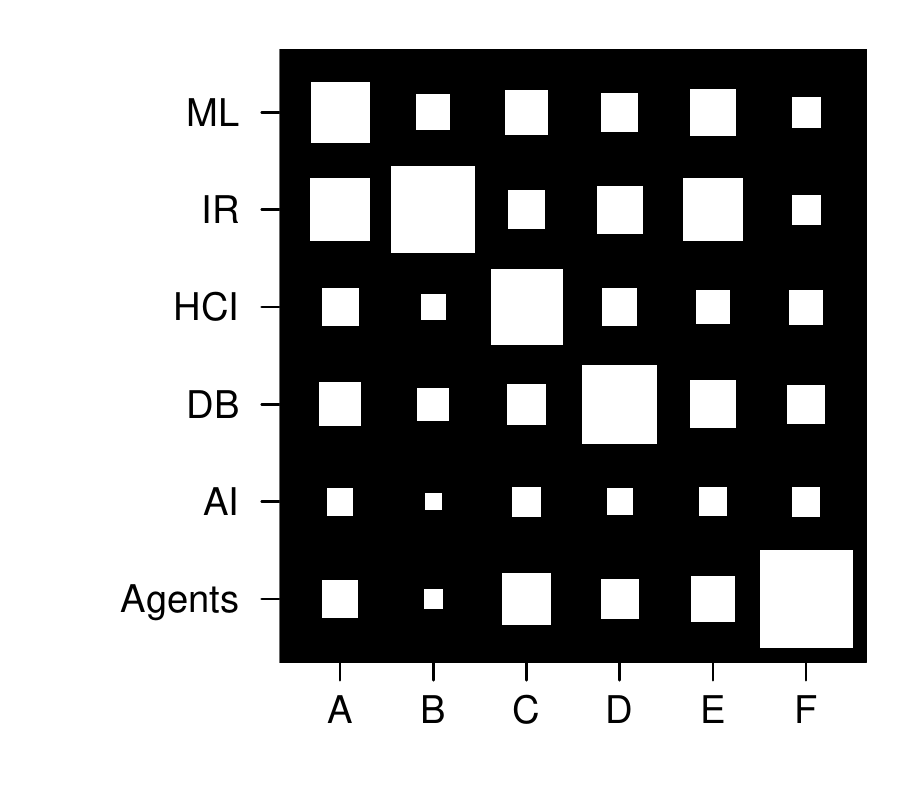}
\end{minipage}
\caption{The models ICMc and SSN-LDA on two citation networks, Cora
  and CiteSeer. \emph{Above:\/} Performance in finding true clusters,
  as measured by the perplexity of predicting ground-truth groups with
  the clusters. The average and 95\% confidence intervals for the mean
  are over 50 chains. \emph{Below:\/} Average confusion matrices
  between the found clusters (columns) and the true clusters (rows).}
\label{fig:perp}
\end{figure}

Over the computed 50 chains, there is a good average correspondence
between the found clusters and the original manual clustering of the
data sets (Fig.~\ref{fig:perp}).  In terms of perplexity ICMc is able
to recover the orignal clusters better than SSN-LDA, although the
average confusion matrices are relatively similar. Results vary from
chain to chain more than with small networks, indicating multiple
local minima for the Gibbs sampler to get trapped into. (See
Section~\ref{sec:d} below for discussion on this behaviour.)

% To quantify the goodness of the clusterings, we constructed an ad hoc
% predictive model from the contingency table. The model gives the
% probability of the real cluster of a node, given its latent component
% in the network model.  That is, the goodness criterion is $\sum_i \log
% p(z_i|c_i)$, with $i$ running over nodes, and the probabilities
% $p(z_i| c_i)$ of the computed clusters given the real clusters
% obtained by normalizing the contingency table. The criterion is
% essentially equivalent to perplexity, cross-entropy between real
% clusters and components, and empirical mutual information---the last
% equivalency holds because the margin of real clusters is fixed.

\subsection{ICMc on Last.fm friendship network}
\label{sec:last}

Last.fm is an Internet site that learns the musical taste of its
members on the basis of examples, and then constructs a personalized,
radio-like music feed. The web site also has a richer array of
services, including a possibility to announce friendships with other
users.  The friendships are initiated by a single party but are later
mutual, forming a network with undirected links. Because friends tend
to be similar, communities in the network would be relatively
homogeneous by their musical taste and other characteristics. We use
this similarity within communities to demostrate ICMc components.

%\documentclass[twoside,11pt]{article}
%\usepackage{../jmlr2e}
%\usepackage{amsmath}
%\usepackage{subfigure}
%\begin{document}

\begin{table}
\begin{center}
  \caption[Last.fm networks and modeling parameters.]{
    Last.fm networks and modeling parameters:
    \(I\) is the number of nodes in the
    network, \(L\) is the number of edges, and \(\alphadp\) and
    \(\beta\) are the hyperparameters.}
\label{tab:lastfmparams}
\smallskip

\begin{tabular}{lll|ll}
&&& \multicolumn{2}{l}{ICMc (DP)}\\
Network & \(I\) & \(L\) &\(\alphadp\)&\(\beta\)\\
\hline
Full Last.fm & 675 682 & 1 898 960 & 0.3 & 0.3 \\
Last.fm USA & 147 610 & 352 987 & 0.2 & 0.2 \\
\end{tabular}

\end{center}
\end{table}

%\end{document} 

\begin{figure}[t]
  \includegraphics[width=\textwidth]{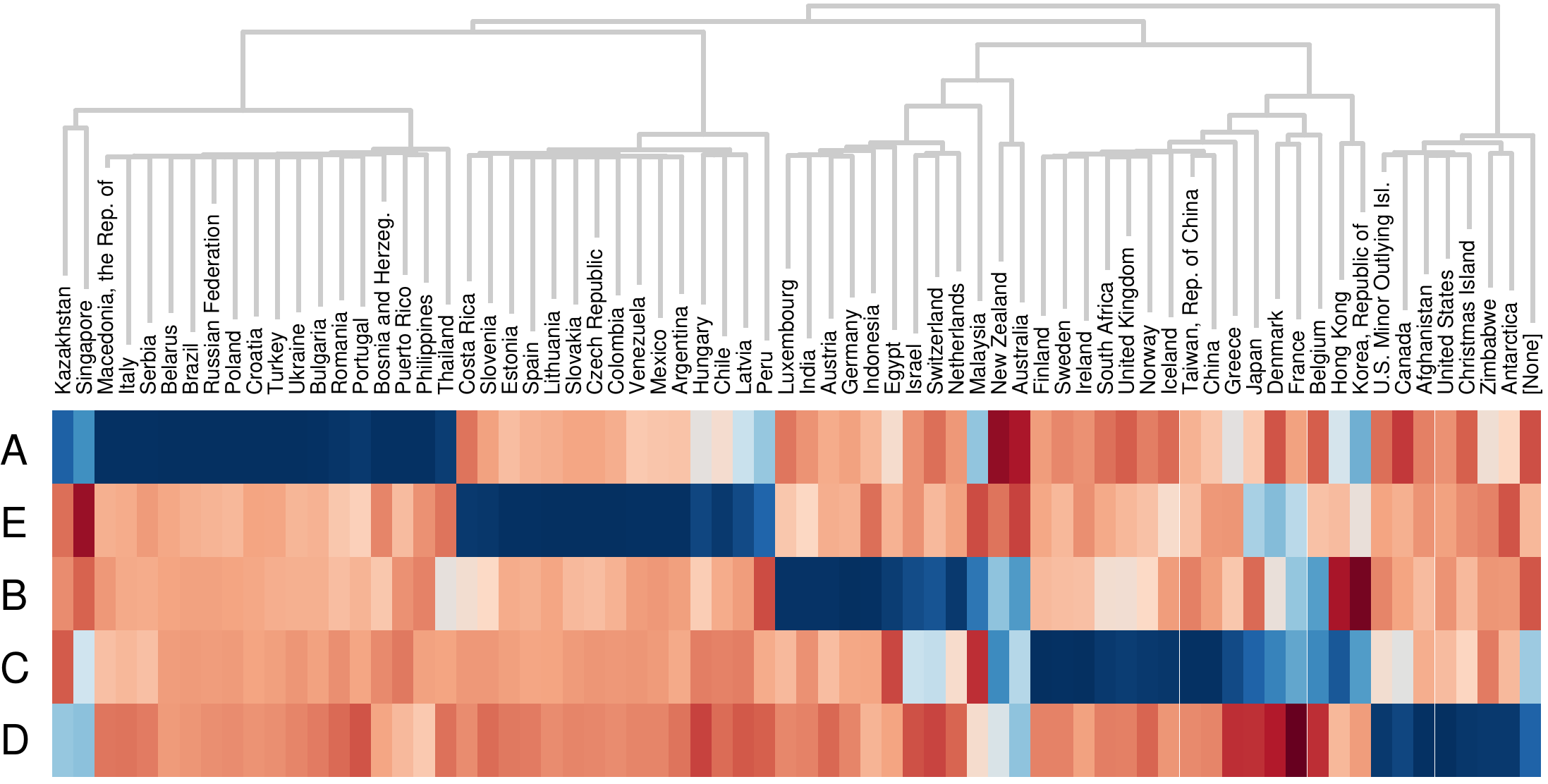}
  \caption{ICMc components (rows) of the Last.fm friendship network
    correlated with nationalities of the participants (columns). The
    full Last.fm network, about 675,000 nodes, was analyzed with the
    DP-version of ICMc.  After a burn-in period of 19,000 iterations,
    20 samples were taken at intervals of 50 iterations to get
    component memberships for the nodes. The running time was 16.4
    hours when run in a single thread on an eight-core 2 GHz Intel
    Xeon. Dark blue and dark red denote the extremes of high and low
    co-occurrence counts, respectively. The columns are ordered and
    the tree produced by \emph{heatmap\/} of the statistical
    environment \emph{R\/}.}
\label{fig:lastfmfull}
\end{figure}

\begin{figure}[t]
  \includegraphics[width=\textwidth]{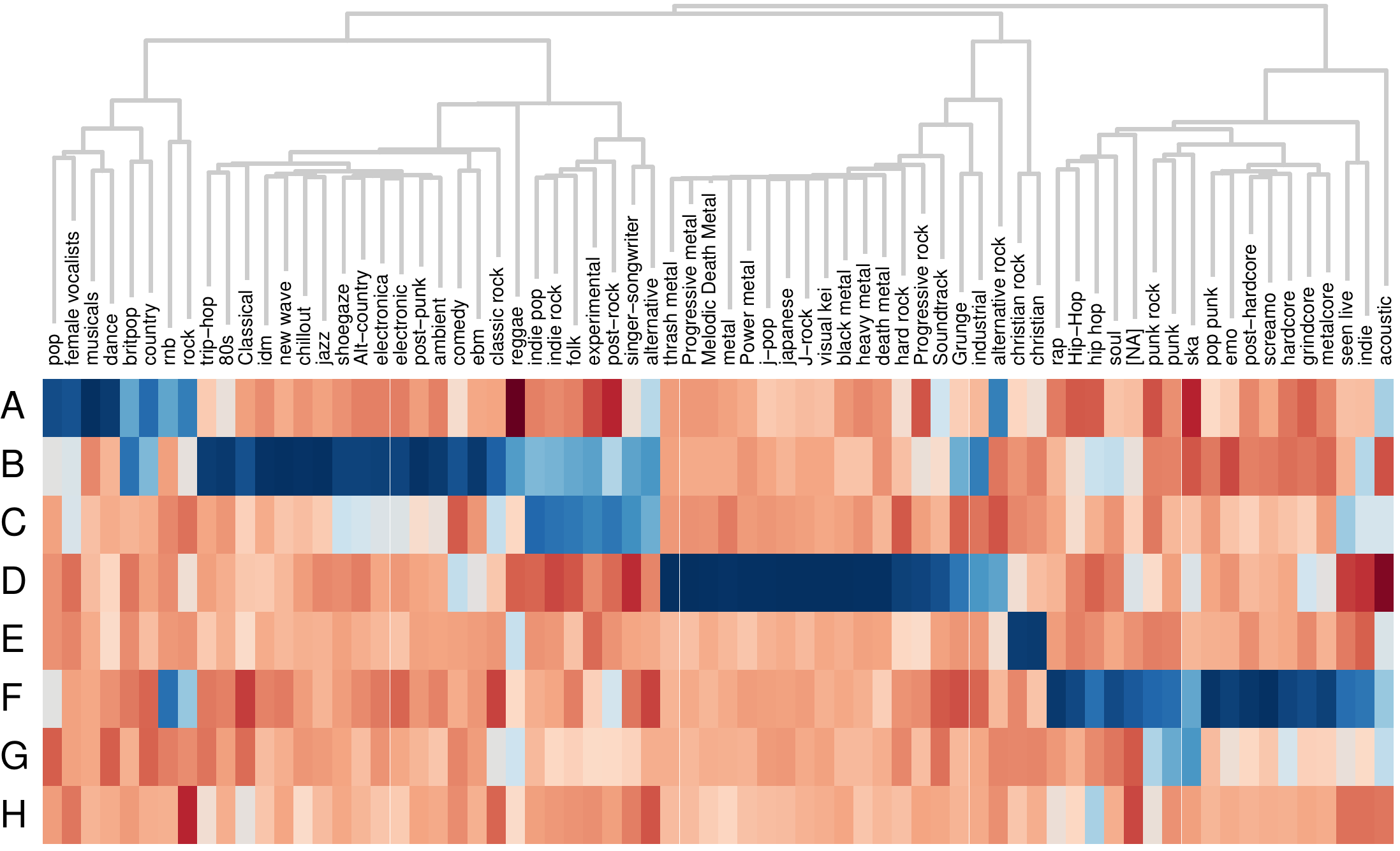}
  \caption{ICMc components (rows) of the Last.fm USA friendship
    network correlated with musical taste of the participants
    (columns).  The model had the DP prior and was computed with a
    burn-in of 49,000 iterations, after which 20 samples were recorded
    at intervals of 50 iterations. The running time was 8.4 hours when
    run in a single thread on an eight-core 2 GHz Intel Xeon. Columns
    correspond to the most common tags given to the songs by the users
    themselves. Other details are as in Fig.~\ref{fig:lastfmfull}.}
\label{fig:lastfmusa}
\end{figure}

The global Last.fm network had about 675,000 nodes and 1,9 million
links, while the subset of US members had about 147,000 nodes and 353,000
links (Table~\ref{tab:lastfmparams}). In addition to the friendships,
we also crawled the nationalities of the site members in the network,
as well as the tags they had associated to the music they
like. The most common tags represent musical genres or subgenres,
allowing interpretation of the components found from the network. 

We modeled the networks with the ICMc, with its Dirichlet process
prior adjusted to favor few components. (With different
hyperparameters, it would have been possible to obtain thousands of
local communities, but the interpretation of such a solution here to
get an idea about its quality would be difficult.) See
Table~\ref{tab:lastfmparams} and Figs.~\ref{fig:lastfmfull}
and~\ref{fig:lastfmusa} for details and results.

The component structure of the full Last.fm network is primarily about
geography or nationalities (Fig.~\ref{fig:lastfmfull}). This was
unexpected at first sight, but in hindsight it is not at all
surprising, for people tend to bond mostly within their country or
city, and the friendships in Last.fm are likely to reflect the
relationships of the real world. Even if they did not, nationality
would affect bonding.  We also correlated the global component
structure to musical taste, and while there are meaningful groups of
genres (not shown, but see Fig.~\ref{fig:lastfmusa}), it is hard to
say which part of them arises due to the geographical division.

Although a more complex model would be needed to find both musical and
geographical structure, the results show that ICMc is able to find
homophilic structures from large networks.  To get a better grasp of
the musical homophily of the network, we also ran ICMc on a
geographically more homogeneous subset of members who have announced
to be from the US. This revealed a clear structure in terms of music
preferences, as shown in Figure~\ref{fig:lastfmusa} and in
Table~\ref{tab:toptags}. The model was able to separate light pop,
more experimental music, ``alternative,'' metal, Christian, and a
punk--hip-hop continuum. In addition, there were two components that
are harder to interpret.

\begin{table}[tbhp]
  \begin{center}
  {\footnotesize
\subtable[]{
\begin{tabular}{lr}\hline\hline
\multicolumn{1}{l}{Cluster A}&
\multicolumn{1}{c}{}
\\ \hline
juggalo&$ 1.36$\\
pop&$ 1.34$\\
musicals&$ 1.32$\\
%Canadian&$ 1.09$\\
%female vocalists&$ 0.85$\\
%singer-songwriter&$ 0.71$\\
%country&$ 0.64$\\
%Soundtrack&$ 0.37$\\
%rock&$ 0.30$\\
%alternative&$ 0.27$\\
\hline
%post-rock&$-1.34$\\
%doom metal&$-1.41$\\
%death metal&$-1.49$\\
%experimental&$-1.55$\\
%grindcore&$-1.58$\\
%indie pop&$-1.60$\\
%synthpop&$-1.65$\\
%shoegaze&$-1.76$\\
Sludge&$\textcolor{red}{-1.89}$\\
black metal&$\textcolor{red}{-1.98}$\\
\hline
\end{tabular}
}
\hspace{0.05in}
\subtable[]{
\begin{tabular}{lr}\hline\hline
\multicolumn{1}{l}{Cluster B}&
\multicolumn{1}{c}{}
\\ \hline
shoegaze&$ 1.35$\\
Alt-country&$ 1.24$\\
post-punk&$ 1.22$\\
%idm&$ 1.15$\\
%new wave&$ 1.15$\\
%psychedelic&$ 1.09$\\
%indie pop&$ 1.08$\\
%ebm&$ 1.05$\\
%synthpop&$ 1.04$\\
%ambient&$ 1.04$\\
\hline
%chinese&$-1.35$\\
%Korean&$-1.35$\\
%emo&$-1.48$\\
%anime&$-1.48$\\
%musicals&$-1.56$\\
%j-pop&$-1.56$\\
%hardcore&$-1.57$\\
%post-hardcore&$-1.62$\\
screamo&$\textcolor{red}{-1.79}$\\
pop punk&$\textcolor{red}{-3.16}$\\
\hline
\end{tabular}
}
\hspace{0.05in}
\subtable[]{
\begin{tabular}{lr}\hline\hline
\multicolumn{1}{l}{Cluster C}&
\multicolumn{1}{c}{}
\\ \hline
indie&$ 0.46$\\
post-rock&$ 0.30$\\
folk&$ 0.22$\\
\hline
%video game music&$-1.17$\\
%Gothic Metal&$-1.21$\\
%psytrance&$-1.28$\\
%japanese&$-1.31$\\
%funk&$-1.33$\\
%hard rock&$-1.42$\\
%heavy metal&$-1.65$\\
%Stoner Rock&$-1.84$\\
visual kei&$\textcolor{red}{-1.86}$\\
j-pop&$\textcolor{red}{-2.08}$\\
\hline
\end{tabular}
}
\hspace{0.05in}
\subtable[]{
\begin{tabular}{lr}\hline\hline
\multicolumn{1}{l}{Cluster D}&
\multicolumn{1}{c}{}
\\ \hline
j-pop&$ 1.69$\\
visual kei&$ 1.68$\\
black metal&$ 1.56$\\
%death metal&$ 1.53$\\
%japanese&$ 1.53$\\
%doom metal&$ 1.51$\\
%Gothic Metal&$ 1.35$\\
%heavy metal&$ 1.32$\\
%Progressive metal&$ 1.31$\\
%video game music&$ 1.30$\\
\hline
%new wave&$-0.77$\\
%singer-songwriter&$-0.78$\\
%soul&$-0.83$\\
%podcast&$-0.88$\\
%folk&$-0.92$\\
%Alt-country&$-1.13$\\
%shoegaze&$-1.13$\\
%indie rock&$-1.19$\\
post-punk&$\textcolor{red}{-1.21}$\\
psychedelic&$\textcolor{red}{-1.41}$\\
\hline
\end{tabular}
}
\hspace{0.05in}
\subtable[]{
\begin{tabular}{lr}\hline\hline
\multicolumn{1}{l}{Cluster E}&
\multicolumn{1}{c}{}
\\ \hline
christian&$ 1.53$\\
podcast&$ 1.01$\\
trance&$ 0.87$\\
%new age&$ 0.70$\\
%Grunge&$ 0.43$\\
%Progressive rock&$ 0.23$\\
%rock&$ 0.22$\\
%singer-songwriter&$ 0.19$\\
%alternative&$ 0.10$\\
\hline
%j-pop&$-0.62$\\
%post-punk&$-0.75$\\
%death metal&$-0.90$\\
%psychedelic&$-1.13$\\
%Canadian&$-1.14$\\
%Gothic Metal&$-1.22$\\
%grindcore&$-1.37$\\
%visual kei&$-1.46$\\
shoegaze&$\textcolor{red}{-1.54}$\\
Sludge&$\textcolor{red}{-1.68}$\\
\hline
\end{tabular}
}
\hspace{0.05in}
\subtable[]{
\begin{tabular}{lr}\hline\hline
\multicolumn{1}{l}{Cluster F}&
\multicolumn{1}{c}{}
\\ \hline
rnb&$ 1.33$\\
screamo&$ 1.21$\\
pop punk&$ 1.15$\\
%post-hardcore&$ 1.07$\\
%hardcore&$ 1.05$\\
%soul&$ 0.99$\\
%emo&$ 0.96$\\
%rap&$ 0.77$\\
%metalcore&$ 0.77$\\
%seen live&$ 0.75$\\
\hline
%idm&$-1.46$\\
%german&$-1.47$\\
%doom metal&$-1.48$\\
%Gothic Metal&$-1.49$\\
%80s&$-1.54$\\
%j-pop&$-1.60$\\
%ebm&$-1.64$\\
%post-punk&$-1.66$\\
Korean&$\textcolor{red}{-2.25}$\\
psytrance&$\textcolor{red}{-2.48}$\\
\hline
\end{tabular}
}
\hspace{0.05in}
\subtable[]{
\begin{tabular}{lr}\hline\hline
\multicolumn{1}{l}{Cluster G}&
\multicolumn{1}{c}{}
\\ \hline
Jam&$ 1.35$\\
ska&$ 0.89$\\
hardcore&$ 0.47$\\
%punk&$ 0.40$\\
%indie&$ 0.11$\\
\hline
%80s&$-0.77$\\
%new age&$-1.00$\\
%christian&$-1.02$\\
%japanese&$-1.04$\\
%soul&$-1.10$\\
%electro&$-1.11$\\
%ebm&$-1.18$\\
%rnb&$-1.36$\\
visual kei&$\textcolor{red}{-1.43}$\\
j-pop&$\textcolor{red}{-2.28}$\\
\hline
\end{tabular}
}
\hspace{0.05in}
\subtable[]{
\begin{tabular}{lr}\hline\hline
\multicolumn{1}{l}{Cluster H}&
\multicolumn{1}{c}{}
\\ \hline
latin&$ 1.13$\\
chinese&$ 1.05$\\
psytrance&$ 0.70$\\
%Korean&$ 0.68$\\
%trance&$ 0.51$\\
%acoustic&$ 0.47$\\
%Hip-Hop&$ 0.44$\\
%rap&$ 0.39$\\
%country&$ 0.33$\\
%metalcore&$ 0.21$\\
\hline
%female vocalists&$-0.37$\\
%experimental&$-0.41$\\
%j-pop&$-0.45$\\
%visual kei&$-0.46$\\
%new wave&$-0.49$\\
%death metal&$-0.51$\\
%post-punk&$-0.56$\\
%Alt-country&$-0.58$\\
synthpop&$\textcolor{red}{-1.54}$\\
juggalo&$\textcolor{red}{-1.62}$\\
\hline
\end{tabular}
}
}
 \end{center}
 \caption{
   The most likely and unlikely tags for each of the ICMc components in the
   Last.fm US network. The tables have been obtained by
   comparing the frequency of the tag to that expected in terms of 
   its marginal probabilities. The table includes only tags for which the
   deviation from the expectation was reliable in terms of a binomial
   test (p=0.05). The numerical values are log-odds.
 }
  \label{tab:toptags}
\end{table}

\section{Discussion}
\label{sec:d}
We have presented two generative models for networks, of which ICMc is
novel, and demonstrated and tested them on data sets of various
sizes. Performance differences between the models were small; ICMc
performed slightly better on networks with strong subgroup cohesion,
while SSN-LDA had an edge in finding more disassortative network
structures. If one is after communities in a social network, there are
both theoretical and empirical reasons to prefer ICMc. The models do
not have significant differences in implementation complexity or ease
of use.

SSN-LDA can be seen as a further development of similar kinds of
models earlier applied to text documents. On the other hand, it is
also a generalization of the model by \citet{Newmanleicht06}, which
interestingly shares notable similarity with earlier text document
models \citep{Hofmann01}. ICMc belongs to the same model family with
LDA, but introduces a generative process that is more faithful to the
idea of subgroup cohesion. An earlier formulation of subgroup cohesion
is modularity \citep{Newmangirvan04,Newman06a}, for which ICMc or its likelihood
could be seen as an alternative. It would be interesting to explore
the relationship between these two, especially as our simulations show
that in general modularity increases monotonically during a Gibbs run
or saturates and only slightly decreases before convergence
\citep{Aukia07}.

Most of our tests were on networks with a known community structure,
which allowed us to set the number of components in advance and use
the Dirichlet prior. In preliminary tests we also tried Dirichlet
process priors with these networks, but performance was naturally
worse since they did not have the prior knowledge about the expected
(``known'') component number. Another reason for the worse performance
of DPs probably is that the size distribution of the communities is
artificially even, for two related reasons.  First, the networks have
survived the selection process of becoming de factor standards for
model testing. Second, in many cases the communities have been
manually set up to make them maximally informative or otherwise
handy. A small number of even-sized communities does not fit well with
the Dirichlet process prior, which assumes either a small number of
communities with rather unequal size, or a very large number with more
equal size. It is likely that in real applications to social and
biological networks the Dirichlet process performs relatively much
better, because real-life communities tends to be of heterogenerous
sizes.

The generative processes of simple models as discussed here are not
meant to be realistic, at least not on higher hierarchical levels
beyond the distributions generating the observed data. Instead, the
ultimate criterion for generative processes should be empirical. Some
abstract information about the networks can be coded to the generative
processes, however. One obvious example is the assortative vs.
disassortative nature of the network structure. It seems that getting
this wrong is not catastrophical, but certainly using the right model
improves performance. Another interesting detail are the Dirichlet
priors. From their urn representations, it is obvious that they mimic
the preferential attachment model of network generation
\citep{Albert02} which produces relatively realistic degree
distributions for social networks.

Even with the Dirichlet process prior one needs to choose the
hyperparameters. Fortunately, the models seem to be quite robust in
terms of the parameter $\beta$, and also in terms of $\alphadir$ with the
Dirichlet prior. It is possible to take $\beta$ into the sampling
process as an MCMC step, because the marginal likelihood for $\beta$
is easy to compute. With the Dirichlet process prior, the parameter
$\alphadp$ fundamentally affects the latent component diversity and
therefore model complexity. For $\alpha$ one can use the proposed
approximations of evidence, such as the harmonic mean estimator
\citep{Griffiths04,Buntine04uai}---it is known to be unstable but at least
sometimes repairable \citep{Raftery07}. Cross validation on the
link level is still another possibility.

Although Gibbs sampling has a reputation of being slow compared to
variational methods, a lot depends on how the slowness is
measured. With topic models for texts, Gibbs is know to produce better
results than variational LDA, at the cost of maybe 4--8 times the
running time to convergence (Wray Buntine, p.c.). But according to
\citet{Griffiths04}, Fig.~1, collapsed Gibbs is actually
\emph{faster}, measured in floating point operations per second to
attain a certain level of perplexity. The difference may partly be
explained by implementational details, but one should also note that
performance measurements should be relative to the goal: While in
statistical inference convergence is essential, in predictive tasks
the predictive performance counts, and often in practice a model is
better if it gives better performance in a shorter running time,
regardless of whether it has converged or not.

In fact, the whole notion of posterior convergence is problematic in
models like LDA and ICMc with a high number of data, parameters and
components. We do know that permutation modes exist and that the
current Gibbs samplers fortunately find only one of them---if they
found more, we would have a label switching problem. Even within a
permutation mode there are probably many local modes of which the
Gibbs sampler explores only part---this is suggested by the variation
between the chains, and the NP-hardness of related formulations of the
community finding problem \citep{Brandes06}. If needed, different
types of compromizes between running time and performance are
available by applying better MCMC techniques, such as annealing,
population methods, or split-merge moves. Variational methods are
available for the DP prior \citep{Blei04} but they are likely to need
help with mode finding.

ICMc and SSN-LDA can be considered as examples of a larger family of
component models, giving generalizations. Links or higher-order
co-occurrences of potentially several types are generated from latent
components, together with other nominal data associated to
nodes. Optimization of such models with collapsed Gibbs is relatively
straightforward and easy to implement, as long as the priors are
conjugate, non-parametric or not. An interesting extension of ICMc,
evidently needed for the Last.fm network, would be to allow factorial
(nominal) components, whose interactions describe the observed
communities. In the Last.fm network, the obvious factors could be
geography and musical taste.  More generic formal extensibility of the
model family, along the lines of relational models \citep[e.g.][]{Xu07} should
also be investigated.

\acks{We thank Last.fm for the data, and E. Airoldi for information on
  the block models, particularly on their scalability. This work was
  supported by Academy of Finland, grant number 119342. SK belongs to the
  Finnish CoE on Adaptive Informatics Research Centre of the Academy
  of Finland and Helsinki Institute for Information Technology, and
  was partially supported by EU Network of Excellence PASCAL.}

\appendix
\section*{Appendix A. The joint distribution and collapsed Gibbs sampler.}
\label{app:deriv}

In ICMc the joint likelihood of observed links $\lset$ and latent
variables $\zset$, given mid-level model parameters $\theta$ and $m$,
is
\[
p(\lset, \zset|m, \theta) = \prod_l \theta_{\zset_l}
m_{\zset_l\,\iset_l} 
m_{\zset_l\,\jset_l} 
= \prod_{z} \theta_z^{n_z} \times \prod_{iz} m^{k_{zi}}_{zi} \;,
\]
where the notation $\zset_l$ refers to the index of the component
generating link $l$, and $\iset_l$ and $\jset_l$ refer to link
endpoint node indices. In the last expression we have link endpoints
counts $n_z$ over components, and $k_{zi}$ over component--node
co-occurrences.  With symmetric Dirichlet priors
$\Dir_\mathrm{sym}^I(\beta)$ for each $m_z$ and
$\Dir_\mathrm{sym}^K(\alphadir)$ for $\theta$, this becomes
\[
p(\lset, \zset, m, \theta|\alphadir, \beta) 
= Z^{-1}(\alphadir, \beta) \prod_{z} \theta_z^{n_z+\alphadir-1} 
\prod_{iz} m^{k_{zi}+\beta-1}_{zi} \;,
\]
with the normalizer $Z$ arising from the Dirichlet priors. Following
\citet{Griffiths04} on Rao-Blackwellisation of LDA,
marginalize over $\theta$ and all $m_z$:
\begin{multline}
p(\lset,\zset|\alphadir, \beta) = \iint p(\lset, \zset,
m,\theta|\alphadir,\beta) \, d\theta\, dm
\\
= 
Z^{-1}(\alphadir, \beta)
\prod_z \frac{\prod_i \g(k_{zi}+\beta)}{\g(2n_z+M\beta)}
\times \frac{\prod_z \g(n_z+\alphadir)}{\g(N+K\alphadir)}\;,
\label{eq:lz}
\end{multline}
where $M$ is the number of nodes, K is the number of components, and
the $2n_z$ comes from the number of component-wise links and the fact
that each link has two endpoints.  (For evaluating the integral, look
for a correspondence with the general Dirichlet distribution and its
normalizing factor.)

Because links are generated independently, they can in principle be
separated from $p(\lset,\zset|\alpha, \beta)$ into link-wise
factors. Separate one arbitrary link, say $l_0$, associated to the
latent variable $z_0$ and to nodes $i_0$ and $j_0$ ($i_0\neq j_0$),
from the product, and denote by $(\lset',\zset')$ the other links
and their associated latent components, and by $(k', n', N')$ the
counts as they were if the link was nonexistent. For most indices, we
will have $k'=k$ and $n'=n$, and always $N'=N-1$, but for some indices
$k'=k-1$ and $n'=n-1$. Because
\begin{eqnarray*}
\g(x) &=& (x-1)\g(x-1)\\
\g(x) &=& (x-1)(x-2)\g(x-2)\;,
\end{eqnarray*}
all this translates into
\begin{multline*}
  p(\lset', \zset', l_0, z_0|\alphadir,\beta) = Z^{-1}(\alphadir, \beta) \prod_z
  \frac{\prod_i \g(k'_{zi}+\beta)}{\g(2n'_z+M\beta)}\times
  \frac{\prod_z \g(n'_z+\alphadir)}{\g(N'+K\alphadir)} \times u_0 \\
= p(\lset',
  \zset'|\alphadir, \beta) \times u_z \; , 
\end{multline*}
where
\begin{equation}
  u_z \equiv p(l_0, z_0|\lset', \zset', \alphadir, \beta) = 
  \frac{(k'_{z_0i_0}+\beta)(k'_{z_0j_0}+\beta)}{(2n'_{z_0}+1+M\beta)(2n'_{z_0}+M\beta)} 
  \times \frac{n'_{z_0}+\alphadir}{N'+K\alphadir}\;.
\label{eq:u0}
\end{equation}
One can use the result to sample a new component $z$ for the left-out
link, with the probabilities $p(z| l_0, \lset', \zset',\alphadir,\beta) =
u_z/u_{\cdot}$, the denominator using the dot notation for the sum. A
Gibbs iteration follows by leaving one link out at a time, and
sampling a new latent component for it as above.

\paragraph{Dirichlet process prior for components.}
% Three ways to derive:
% - Talk about urns, then Polya to Blackwell-MacQueen. This is least formal.
% - Substitute \alphadp to likelihood, then take a limit
% - Take p(z|alpha) from some source, and/or refer to Blackwell-MacQueen, then marginalize m
The ICMc model can be derived for a Dirichlet Process component prior
in several ways. Informally, after seeing the link removal
decomposition with $u_z$, one notes the structure of
$p(\lset,\zset|\alphadir,\beta)$ as nested Polya urns \citep{Johnson77}. One can
then substitute the component urn, the last factor in (\ref{eq:u0}),
with the Blackwell-MacQueen urn \citep{Blackwell73,Tavare97} parameterized by $\alphadp$:
\begin{equation}
  p(l_0, z_0|\lset',\zset',\alphadp,\beta) = 
  \frac{(k'_{z_0i_0}+\beta)(k'_{z_0j_0}+\beta)}{(2n'_{z_0}+1+M\beta)(2n'_{z_0}+M\beta)} 
  \frac{\C(n_{z_0}, \alphadp)}{N+\alphadp}
\label{eq:u0dp}
\end{equation}
with $\C(n,\alpha)=n$ if $n \neq 0$ and $\C(0,\alpha)=\alpha$.

Another way to end up with the same result is to substitute
$\alphadir=\alphadp/K$ to (\ref{eq:lz}) or (\ref{eq:u0}), then collect
all empty components into one bin, and take the limit $K\to\infty$
\citep{Neal00}.

More formally, one can first write the joint distribution of the ICMc model
with an unspecified component prior $p(\zset|\alpha)$,
\[
p(\lset, \zset, m|\alpha, \beta) = p(\lset,m | \zset,\beta)\, p(\zset|\alpha) =
Z^{-1}(\beta)
%\prod_{z} \theta_z^{n_z+\alpha-1} % Component prior for the Dirichlet. don't loose.
\prod_{iz} m^{k_{zi}+\beta-1}_{zi} 
\times
p(\zset|\alpha)
 \;,
\]
integrate $m$ out, and then substitute the Dirichlet process prior
(e.g., Dahl 2003), obtainable from the Blackwell-Queen urn model by
induction, to end up with
\begin{equation}
  p(\lset, \zset|\alphadp, \beta)
  =
  Z^{-1}(\beta)\, 
  \prod_z \frac{\prod_i \g(k_{zi}+\beta)}{\g(2n_z+M\beta)} 
  \times
  \frac{\alphadir^K\, \Gamma (\alphadir)}{\Gamma (\alphadir+N)} 
  \prod_z \mathrm{\Gamma}(n_z) \;.
\label{eq:lzdp}
\end{equation}
The sampling rule (\ref{eq:u0dp}) can then be obtained by computing
the probability of one (removed) link given all others, just as in the
case of a finite Dirichlet prior.

\paragraph{Collapsed Gibbs sampling for the SSN-LDA model.}
The collapsed sampler is identical to that in \citet{Griffiths04},
also presented by \citet{Zhang07isi}. The collapsed sampling formula
for SSN-LDA with the DP prior is obtained analogously to ICMc, by
modifying the factor corresponding to the latent-component urn.

\vskip 0.2in
% Siirretään ja uudelleennimetään kun on saatu tää papru tehtyä
% TODO: Korjaa isot alkukirjaimet .bib-tiedostosta
\bibliography{network-jannea}

\end{document}